\documentclass[runningheads]{llncs}

\usepackage{eccv}

\usepackage{eccvabbrv}

\usepackage{graphicx}
\usepackage{booktabs}

\usepackage[accsupp]{axessibility}  

\usepackage{multirow}
\usepackage{tabularx,booktabs}
\usepackage{adjustbox}
\newcolumntype{C}[1]{>{\centering\arraybackslash}p{#1}}
\usepackage{amsmath, bm}
\usepackage{subcaption}
\usepackage{xcolor,colortbl}
\definecolor{LightCyan}{rgb}{0.88,1,1}

\usepackage[pagebackref,breaklinks,colorlinks,citecolor=eccvblue]{hyperref}

\usepackage{orcidlink}

\begin{document}

\title{PatchRefiner: Leveraging Synthetic Data for Real-Domain High-Resolution Monocular Metric Depth Estimation} 

\titlerunning{PatchRefiner}

\author{Zhenyu Li, Shariq Farooq Bhat, Peter Wonka}


\institute{King Abdullah University of Science and Technology (KAUST)\\
\small\url{https://github.com/zhyever/PatchRefiner}\\
{\tt\small zhenyu.li.1@kaust.edu.sa}}

\maketitle

\begin{abstract}
  
  This paper introduces PatchRefiner, an advanced framework for metric single image depth estimation aimed at high-resolution real-domain inputs. While depth estimation is crucial for applications such as autonomous driving, 3D generative modeling, and 3D reconstruction, achieving accurate high-resolution depth in real-world scenarios is challenging due to the constraints of existing architectures and the scarcity of detailed real-world depth data.
  PatchRefiner adopts a tile-based methodology, reconceptualizing high-resolution depth estimation as a refinement process, which results in notable performance enhancements.
  Utilizing a pseudo-labeling strategy that leverages synthetic data, PatchRefiner incorporates a Detail and Scale Disentangling (DSD) loss to enhance detail capture while maintaining scale accuracy, thus facilitating the effective transfer of knowledge from synthetic to real-world data.
  Our extensive evaluations demonstrate PatchRefiner's superior performance, significantly outperforming existing benchmarks on the Unreal4KStereo dataset by 18.1\% in terms of the root mean squared error (RMSE) and showing marked improvements in detail accuracy and consistent scale estimation on diverse real-world datasets like CityScape, ScanNet++, and ETH3D. 
\end{abstract}

\section{Introduction}
\label{sec:intro}

\begin{figure}[t]
    \centering
    \begin{subfigure}{0.19\textwidth}
        \centering
        \includegraphics[width=1\linewidth]{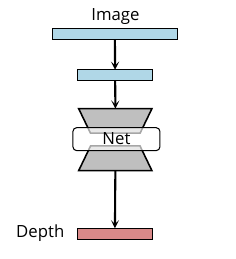}
        \caption{LR Depth}
        \label{fig:figsub:f1}
    \end{subfigure}
    \hfill
    \begin{subfigure}{0.39\textwidth}
        \centering
        \includegraphics[width=1\linewidth]{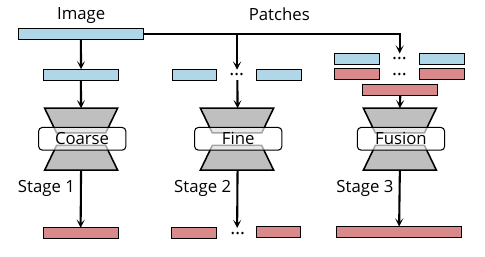}
        \caption{Fusion-Based Framework}
        \label{fig:figsub:f2}
    \end{subfigure}
    \begin{subfigure}{0.39\textwidth}
        \centering
        \includegraphics[width=1\linewidth]{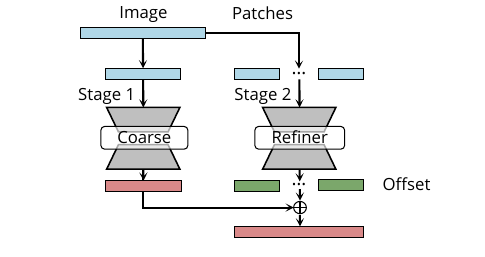}
        \caption{Refiner-Based Framework}
        \label{fig:figsub:f3}
    \end{subfigure}
    \caption{\textbf{Framework Comparison.} (a) Low resolution depth estimation framework with single forward pass. (b) Fusion-based high-resolution framework combining the best of coarse and fine depth predictions~\cite{li2023patchfusion,miangoleh2021boostingdepth}. (c) Our refiner-based framework predicts a residual to refine the coarse prediction.}
    \label{fig:framework}
\end{figure}

This paper delves into the field of metric single-image depth estimation, focusing on high-resolution inputs from the real domain. High-resolution depth estimation plays a key role in autonomous driving, augmented reality, content creation, and 3D reconstruction~\cite{eigen2014mde,bhat2023zoedepth,li2022binsformer,zhang2023controlnet}. Despite significant progress, the high-resolution depth estimation in real-world scenarios remains daunting. This challenge is primarily due to the resolution limitations inherent in most state-of-the-art depth estimation architectures~\cite{bhat2023zoedepth,yang2024depthanything,zhao2023unleashing} and the scarcity of high-quality real-world depth datasets.

The current state-of-the-art in high-resolution depth estimation, PatchFusion~\cite{li2023patchfusion,miangoleh2021boostingdepth,rey2022360monodepthtile}, employs a tile-based strategy to navigate the resolution constraints, posing the task as a fusion process of coarse and fine depth estimations. Because of the scarcity of real high-resolution depth datasets, PatchFusion resorts to training on a synthetic 4K dataset~\cite{li2023patchfusion, tosi2021smd}.
There are two limitations of PatchFusion that we would like to improve upon: 1) It utilizes a three-step training process, which is not only time-consuming and expensive but also risks the framework achieving the stage-wise local optima and limits the potential performance gains from end-to-end learning. 2) PatchFusion demonstrates poor generalization to the real domain. 

The poor synthetic-to-real generalization is a long-standing problem in general~\cite{farahani2021brief,weiss2016survey}. In the context of depth estimation, the difference in the metric scale and depth distributions of synthetic datasets and the real domain further exacerbates the domain shift. This results in particularly subpar scale accuracy of depth models on real data when trained on synthetic datasets. On the other hand, the real datasets are not only low resolution but also often have missing ground truth values due to sensor constraints, occlusion, etc (see Fig.~\ref{fig:dataset} and Sec.~\ref{subsec:dsd}). This adds to the inability of depth models to capture sharp details when trained on real datasets. Thus, one is confronted with a dilemma - Training on real datasets leads to good scale accuracy but poor high-frequency details, while training on synthetic datasets leads to sharper results but poor scale performance on real images. 

We introduce \textbf{PatchRefiner}, a novel framework that reformulates high-resolution depth estimation as a process of refining coarse depth. We propose improvements on two levels:

First, unlike direct depth regression approaches~\cite{li2023patchfusion,poucin2021boosting,rey2022360monodepthtile}, PatchRefiner utilizes a frozen coarse depth model and enhances the quality by predicting residual depth for refinement. This approach not only streamlines model training but also markedly enhances performance.

Second, we propose a method to exploit the best of both worlds to solve the above synthetic-real dilemma. 
We employ a teacher-student framework, leveraging the sharpness of synthetic data while learning the scale from real data. The teacher model, pre-trained on synthetic data, generates pseudo labels for real-domain training samples. Recognizing that these pseudo labels offer detailed features albeit with scale inaccuracies, we introduce the Detail and Scale Disentangling (DSD) loss. This loss integrates ranking supervision and scale-shift invariance, drawing inspiration from recent advances in relative depth estimation~\cite{chen2016diw,xian2020structurediw,Ranftl2022midas,yang2024depthanything}. It leads to a framework capable of delivering high-resolution depth estimates with both precise scale and sharp details in real-world settings.

Our evaluation of PatchRefiner on the Unreal4KStereo synthetic dataset~\cite{tosi2021smd} demonstrates a substantial improvement over the current state-of-the-art, reducing RMSE by 18.1\% and REL by 15.7\%. Further, we assess the framework's efficacy in leveraging synthetic data across diverse real-world datasets, including CityScape~\cite{cordts2016cityscapes} (outdoor, stereo), ScanNet++~\cite{yeshwanth2023scannet++} (indoor, LiDAR and reconstruction), and ETH3D~\cite{schops2017eth3d} (mixed, LiDAR). Our findings reveal notable enhancements in depth boundary delineation (e.g., a 19.2\% increase in boundary recall on CityScape) while maintaining accurate scale estimation, showcasing the framework's adaptability and effectiveness across varying settings and sensor technologies.

\section{Related Work}

\noindent \textbf{High-Resolution Monocular Metric Depth Estimation.} Monocular depth estimation has achieved tremendous progress~\cite{eigen2014mde,fu2018dorn,godard2019mde2,lee2020multiloss,bhat2021adabins,li2022binsformer,li2023depthformer}. Current SOTA approaches often employ complex network architectures yet grapple with the limitations imposed by their input resolution~\cite{bhat2023zoedepth,yang2024depthanything}. This stands in stark contrast to the advancements in modern imaging devices that capture images at increasing resolutions, and the growing demand among users for high-resolution depth estimation. Initial efforts to address this gap included the use of Guided Depth Super-Resolution (GDSR)~\cite{zhao2022dgsrdiscrete,metzger2023gdsrdiff,hui2016gdsrdepth,zhong2023guided} and Implicit Functions~\cite{mildenhall2021nerf,chen2021liif}. Recently, Tile-Based Methods have emerged as a potent strategy for high-resolution depth estimation~\cite{li2023patchfusion,miangoleh2021boostingdepth,rey2022360monodepthtile}, segmenting images into patches for individual processing before reassembling them into a comprehensive depth map. This paper extends the tile-based approach, seeking to elevate the quality of depth estimation further.

\noindent \textbf{Synthetic Data for Depth Estimation.} The challenge of acquiring high-quality, real-domain data for high-resolution depth training has prompted the use of synthetic datasets~\cite{rajpal2023high}. Traditionally, synthetic data has been employed within unsupervised domain adaptation frameworks, utilizing labeled synthetic and unlabeled real-domain data to enhance depth estimation accuracy on real-world images~\cite{chen2019crdoco,kundu2018adadepth,lopez2023desc,koutilya2020sharingan,zhao2019geometry,zheng2018t2net}. Techniques vary from pixel-level style transfer and image translation to feature-level adversarial learning~\cite{chen2019crdoco,kundu2018adadepth,zhao2019geometry,zheng2018t2net}, with some methods integrating additional information such as stereo pairs or segmentation maps for enhanced adaptation~\cite{lopez2023desc,koutilya2020sharingan,zhao2019geometry}. Contrasting with these approaches, our work explores a practical scenario where labeled data from both synthetic and real domains are leveraged to improve real-world, high-resolution depth estimation, delving into a relatively underexplored application of synthetic data.


\noindent \textbf{Pseudo-Labeling for Depth Estimation.} Pseudo-labeling, a cornerstone of semi-supervised learning, has been widely applied across various domains, including classification~\cite{chen2019progressive,hu2021simple,pseudo2013simple,saito2017asymmetric,taherkhani2021self} and scene understanding~\cite{li2022unsupervised,chen2020digging,li2019bidirectional,pastore2021closer,paul2020domain,shin2022mm,zhao2020object,zou2018unsupervised}, to extrapolate knowledge from labeled data to unlabeled datasets. In depth estimation, pseudo-labeling often serves to provide supplementary supervision in unsupervised domain adaptation settings~\cite{lopez2023desc,yen20223d,yang2021self}. While recent state-of-the-art methods like Depth-Anything utilize pseudo-labeling to enhance model generalization~\cite{yang2024depthanything}, these techniques predominantly aim to refine or enhance the pseudo labels themselves. Our approach diverges by utilizing real-domain data with accurate depth labels, focusing on a novel Detail and Scale Disentangling loss. This loss mechanism uniquely leverages the detailed insights from pseudo labels to enrich real-domain depth estimation without compromising the scale accuracy derived from real ground-truth data.

\begin{figure}[t]
    \centering
    \includegraphics[width=0.9\linewidth]{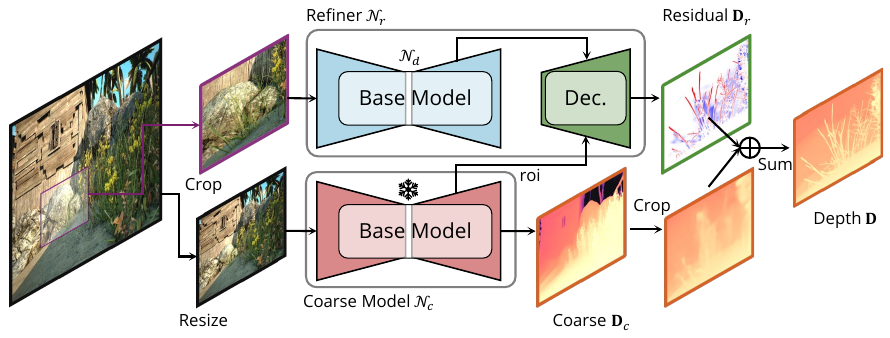}
    \caption{\textbf{Architecture Illustration.} PatchRefiner contains a pre-trained frozen coarse depth estimation model $\mathcal{N}_c$ and a refiner model $\mathcal{N}_r$ predicts residual depth map $\mathcal{D}_r$ to refine the coarse depth $\mathcal{D}_c$. The refiner contains one base depth model $\mathcal{N}_d$ that has the same architecture as $\mathcal{N}_c$, and a light-weight decoder to aggregate information and make the final prediction.}
    \label{fig:arch}
\end{figure}

\section{Method}
\label{sec:method}

In this section, we first present the overall PatchRefiner framework in Sec.~\ref{subsec:framework}. Then, we introduce the limitation of adopting real-domain data for high-resolution depth estimation~\ref{subsec:limit} and the proposed teacher-student framework in Sec.~\ref{subsec:s2rpipeline} with the Detail and Scale Disentangling loss (DSD) in Sec.~\ref{subsec:dsd}.

\subsection{PatchRefiner Framework}
\label{subsec:framework}

PatchRefiner follows the tile-based strategy to address the prohibitive memory and computational demands for high resolutions such as 4K~\cite{li2023patchfusion, miangoleh2021boostingdepth}. However, recognizing the limitations of existing models, we propose a simplified two-step approach for high-resolution depth estimation: \textbf{(i)} Coarse Scale-Aware Estimation, and \textbf{(ii)} Fine-Grained Depth Refinement, shown in Fig.~\ref{fig:arch}.

\textbf{(i) Coarse Scale-Aware Estimation:} The foundation of PatchRefiner is the Coarse Depth Estimation network, $\mathcal{N}_{c}$, which processes downsampled versions of the input images to produce a global depth prediction, $\mathbf{D}_{c}$. Similar to previous work~\cite{li2023patchfusion, miangoleh2021boostingdepth}, this step is crucial for establishing a baseline depth map that captures the scene's overall structure and depth consistency, albeit without high-resolution details. Moreover, $\mathcal{N}_{c}$ can be an arbitrary depth estimation model and it is frozen after the first step of training.

\textbf{(ii) Depth Refinement Process:} Different from the conventional approach of using a separate fine depth network and a fusion mechanism~\cite{li2023patchfusion,poucin2021boosting}, our framework introduces a unified refinement network, $\mathcal{N}_{r}$. This network is designed to refine the coarse depth map by focusing on the recovery of lost details and enhancing depth precision at a patch-wise level. 

The input to $\mathcal{N}_{r}$ is the cropped original image $I$, which is processed by the base depth model $\mathcal{N}_{d}$ with the same architecture as $\mathcal{N}_{c}$ in the refiner module. Then, we collect $L$-level multi-scale immediate features from both $\mathcal{N}_{d}$ and $\mathcal{N}_{c}$, denoting them as $\mathcal{F}_d=\{f_d^i\}_{i=1}^{L}$ and $\mathcal{\Tilde{F}}_c=\{\Tilde{f}_c^i\}_{i=1}^{L}$, respectively. Following~\cite{li2023patchfusion}, we apply the $\texttt{roi}$~\cite{he2017mask} operation to fetch features of the corresponding cropped area as $\Tilde{f}_c^i=\texttt{roi}(f_c^i)$.

Next, we adopt a lightweight decoder to obtain high-resolution predictions.
Given $\mathcal{F}_d$ and $\mathcal{\Tilde{F}}_c$, we aggregate them with concatenation operators ($\texttt{Cat}$) following by simple convolutional blocks ($\texttt{CB}$):

\begin{equation}
\label{eq:fuse}
    f_r^i = \texttt{CB}(\texttt{Cat}(f_d^i, \Tilde{f}_c^i)),
\end{equation}

The output feature set $\mathcal{F}_r=\{f_r^i\}_{i=1}^{L}$ is then fed to a successive series of up-sampling layers~\cite{lehtinen2018noise2noise}, in order to construct the residual depth map $\mathbf{D}_{r}$ at the input resolution. Associated with skip-connections, they form our refiner decoder. Further details about the architecture with immediate feature shapes are described in the supplementary material. Finally, we calculate the final patch-wise depth estimation as $\mathbf{D} = \texttt{roi}(\mathbf{D}_{c}) + \mathbf{D}_{r}$.

\subsection{Limitation of Real-Domain Depth Estimation Datasets}
\label{subsec:limit}

Our second goal in this paper is to train a real-domain high-resolution depth estimation model with both synthetic dataset $\mathcal{S}$ and real-domain dataset $\mathcal{R}$. Our main insight is to distinguish between \emph{scale errors} and \emph{boundary errors}.

In the field of high-resolution depth estimation, the prevailing state-of-the-art methodology~\cite{li2023patchfusion} trains models using synthetic datasets, which offers paired high-resolution images and corresponding dense high-resolution depth ground-truth maps~\cite{tosi2021smd,huangDeepMVS2018}. This synthetic training regime, while beneficial in a controlled setting, introduces significant challenges when models are applied to real-world data due to the intrinsic domain gap between synthetic and real-world environments. This gap often manifests as substantial scale errors during inference on real-domain datasets as shown in Tab.~\ref{tab:s2r}.

\begin{figure}[t]
    \centering
    \includegraphics[width=1\linewidth]{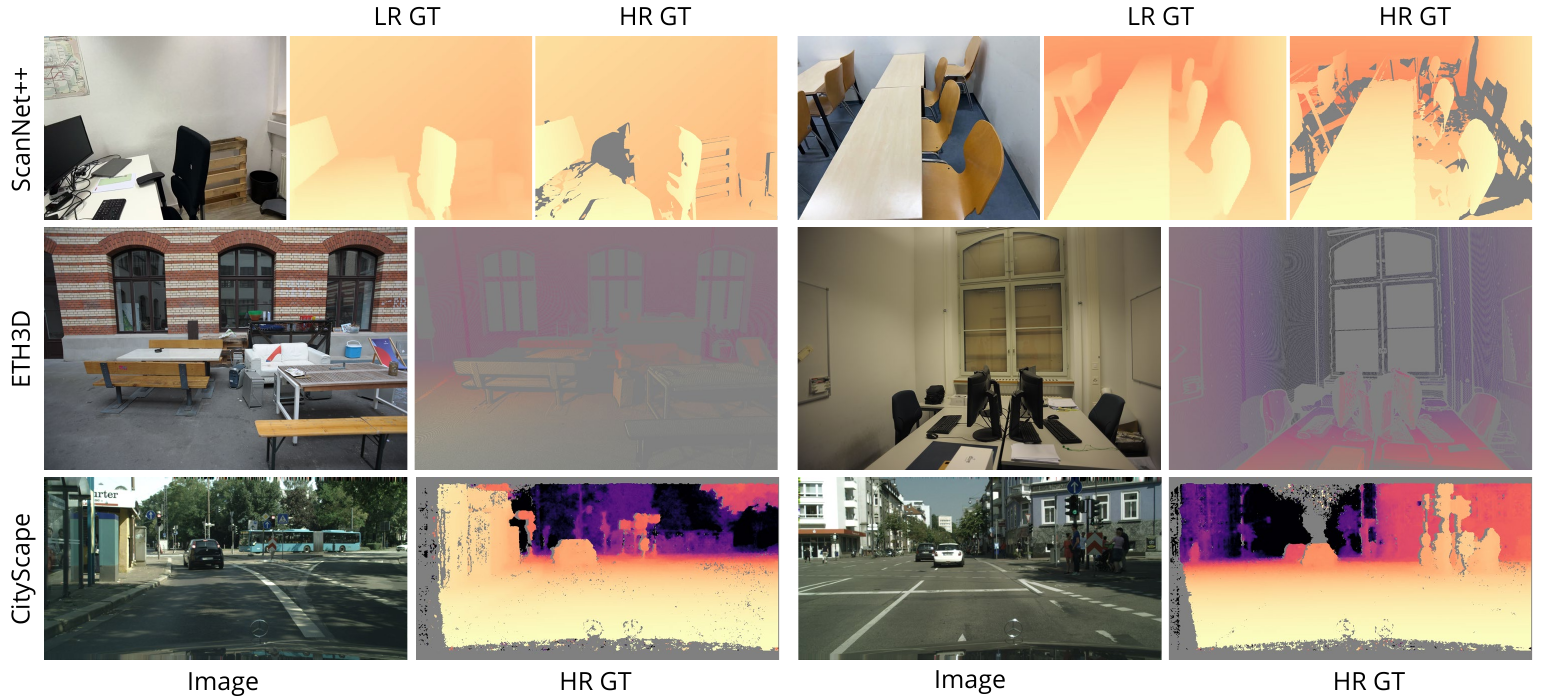}
    \caption{\textbf{Visualization of Real-Domain Data Pairs.} Points lacking ground-truth data are depicted in gray. Due to sparse annotations near edges, models trained on real-domain data exhibit blurred boundary estimations.}
    \label{fig:dataset}
\end{figure}

Addressing this issue directly by training on real-domain datasets presents its own set of challenges. Real-world high-resolution depth datasets are scarce and typically constrained by limited resolution and unavailable missing ground-truth pixels. These limitations stem from the methods employed in generating real-world depth annotations, such as Kinect~\cite{silberman2012nyu,janoch2013categorydata,song2015sunrgbd}, LiDAR~\cite{yeshwanth2023scannet++,schops2017eth3d,geiger2012kitti}, or stereo vision techniques~\cite{cordts2016cityscapes,scharstein2014mid}, each with their inherent drawbacks as shown in Fig.~\ref{fig:dataset}.

For instance, depth maps generated using Kinect technology are confined to a resolution of 640×480~\cite{silberman2012nyu,song2015sunrgbd}, insufficient for high-resolution depth estimation~\cite{li2023patchfusion}. LiDAR-based depth maps, while useful, tend to be sparse~\cite{geiger2012kitti,schops2017eth3d} or limited in resolution (e.g., 256x192 in Scannet++~\cite{yeshwanth2023scannet++}), and the process of reconstructing high-resolution dense depth maps from LiDAR point clouds is fragile with cascade errors and omissions~\cite{yeshwanth2023scannet++}. Stereo vision techniques, on the other hand, can also introduce missing values around object boundaries due to rectification transformations~\cite{cordts2016cityscapes,scharstein2014mid}. It is important to realize that existing high-resolution real datasets with missing values around edges do not help in reducing \emph{boundary errors}.

These limitations highlight the inherent difficulties in employing real-world datasets for training high-resolution depth estimators~\cite{li2023patchfusion,poucin2021boosting}. The lack of high-quality, high-resolution ground-truth depth maps in the real domain makes it challenging to train models that can accurately predict sharp depth around fine object boundaries~\cite{rajpal2023high}. 
Our main idea is to devise a training strategy that can improve the \emph{scale errors} when fine-tuning on real data, while maintaining high-resolution information around edges to minimize \emph{boundary errors}.

\begin{figure}[t]
    \centering
    \includegraphics[width=0.95\linewidth]{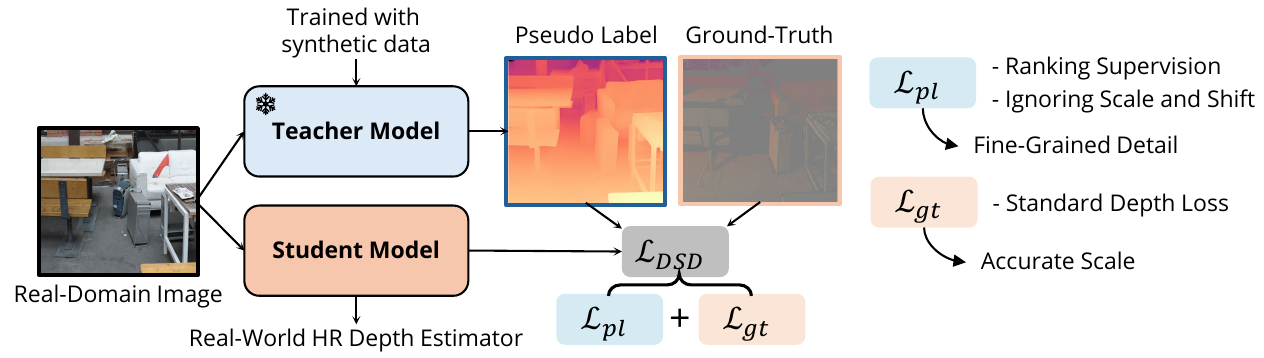}
    \caption{\textbf{Enhancing Real-Domain Learning with Synthetic Data}. A teacher model trained on synthetic data produces pseudo labels for real-domain training. The student model benefits from a DSD dual-supervision approach: loss on pseudo labels for detail enhancement and loss on ground truth for scale accuracy. This method ensures detailed depth perception without compromising scale accuracy.}
    \label{fig:s2r}
\end{figure}

\subsection{Overall Pipeline Illustration}
\label{subsec:s2rpipeline}

Building on the success of semi-supervised learning frameworks in depth estimation tasks~\cite{yang2024depthanything,yen20223d,lopez2023desc}, our proposed framework employs a teacher-student architecture, as shown in Fig.~\ref{fig:s2r}, to effectively integrate synthetic and real-domain data for high-resolution depth estimation. It employs a two-step process:

\textbf{Teacher Model Training:} The teacher model is initially pretrained on a synthetic dataset $\mathcal{S}$, which, due to its high-quality and detailed annotations, enables the model to predict high-resolution depth maps with precise boundaries. 

\textbf{Student Model Training with both Pseudo Labels and Ground-Truth Depth:} In the subsequent phase, the teacher model is frozen, and the student model is trained on the real-domain dataset $\mathcal{R}$, utilizing both the ground truth depth labels $\Tilde{\mathbf{D}}$ and pseudo labels $\hat{\mathbf{D}}$ generated by the teacher model. As identified in the limitations of real-domain datasets, while the ground truth labels offer accurate depth information, they miss crucial information near boundaries essential for high-resolution depth learning. To address this, the student model is guided by the teacher's pseudo labels, which excel near boundaries. However, these pseudo labels, while sharp, exhibit scale discrepancies due to the domain gap between synthetic and real-world data as shown in Tab.~\ref{tab:s2r}.

Therefore, we introduce the Detail and Scale Disentangling (DSD) loss. It ensures that the final high-resolution depth predictions from the student model maintain an accurate scale while benefiting from the enhanced boundary details provided by the teacher model trained on the synthetic data.

\subsection{Detail and Scale Disentangling Loss}
\label{subsec:dsd}

The student model is subject to two sources of supervision. The primary one is the standard scale-invariant loss~\cite{eigen2014mde,bhat2023zoedepth,li2022binsformer,li2023patchfusion}, $\mathcal{L}_{silog}$, calculated with the ground truth (GT) data, which ensures reliable and scale-consistent depth estimation on the real-domain data.

The secondary source of supervision comes from pseudo labels generated by the teacher model. These labels, while detailed in capturing boundaries and fine details, exhibit low scale accuracy. Directly applying conventional depth losses~\cite{eigen2014mde, lee2020multiloss,liu2023va} could lead to an imbalance between enhancing detail and maintaining scale accuracy. To address this, motivated by current impressive progress in relative depth estimation~\cite{chen2016diw,xian2020structurediw,Ranftl2022midas,yang2024depthanything}, we adopt the ranking loss $\mathcal{L}_{rank}$~\cite{chen2016diw,xian2020structurediw} and the Scale and Shift Invariant losses $\mathcal{L}_{ssi}$~\cite{Ranftl2022midas,yang2024depthanything} to inject the detail information. Notably, The $\mathcal{L}_{rank}$ only provides supervision on the prediction relationship whereas $\mathcal{L}_{ssi}$ ignores scale and shift. Both losses tackle the same challenge with a slightly different approach.

\noindent\textbf{Ranking loss}~\cite{chen2016diw,xian2020structurediw} is designed for sparse sets with ordinal data. This adaptation allows us to leverage dense detailed ranking information from the pseudo labels, enhancing high-resolution estimation without compromising scale accuracy. Specifically, for a certain pair of points $[p_{i,1}, p_{i,2}]$ with predicted depth values $[d_{i,1}, d_{i,2}]$ and pseudo depth values $[\hat{d}_{i,1}, \hat{d}_{i,2}]$ in a set of $N$ sampled point pairs $\mathcal{P}=\{[p_{i,1}, p_{i, 2}], i=1,2,\cdots,N\}$, the ranking loss is defined as 

\begin{equation}
    \mathcal{L}_{rank} = \frac{1}{N} \sum\limits_{i}\phi(p_{i,1}, p_{i,2}),
\end{equation}

\begin{equation}
  \phi(p_{i,1}, p_{i,2}) =
    \begin{cases}
      \log(1+\exp(-\ell\times (d_{i,1}-d_{i,2}))), & \ell \neq 0, \\
      (d_{i,1}-d_{i,2})^2, & \ell = 0,
    \end{cases}       
\end{equation}

\noindent where $\ell$ is the pseudo ordinal label, which can be induced by:
\begin{equation}
  \ell =
    \begin{cases}
      +1, & \hat{d}_{i,1}/\hat{d}_{i,2} >= 1 + \tau, \\
      -1, & \hat{d}_{i,1}/\hat{d}_{i,2} <= \frac{1}{1+\tau}, \\
      0, & \mathrm{otherwise}.
    \end{cases}       
\end{equation}

\noindent Here $\tau$ is a tolerance threshold, which is set to 0.03 in our experiments following~\cite{xian2020structurediw}. This loss not only encourages the predicted depth values for closely related points to align but also emphasizes the differentiation of points when the pseudo depth value of the two points is different.

\noindent\textbf{Scale and Shift Invariant loss}~\cite{Ranftl2022midas,yang2024depthanything} is proposed to learn relative depth estimation with ignoring the unknown scale and shift of each sample:

\begin{equation}
    \mathcal{L}_{ssi} = \frac{1}{M} \sum\limits_{i=1}^{M}\rho(d_i^*-\hat{d}_i^*),
\end{equation}
\noindent where $d_i^*$ and $\hat{d}_i^*$ are scaled and shifted versions of the predicted depth $d_i$ and pseudo label $\hat{d}_i$, and $\rho$ is the mean absolute error loss. $M$ is the number of pixels. We use the least-squares criterion to align the prediction to the ground truth:
\begin{equation}
    (s,t) = \mathrm{argmin}_{s,t} \sum\limits^M_{i=1} (sd_i + t - \hat{d_i})^2,
\end{equation}
\begin{equation}
    d^* = sd + t,~ \hat{d}^*=\hat{d} 
\end{equation}
\noindent where the scale $s$ and shift $t$ factors are effectively determined with the closed form~\cite{Ranftl2022midas}. 

The final Detail and Scale Disentangling loss for training the student model combines the $\mathcal{L}_{silog}$, $\mathcal{L}_{rank}$, and $\mathcal{L}_{ssi}$ as follows:

\begin{equation}
\label{eq:semi-overall-loss}
     \mathcal{L}_{DSD} = \overbrace{\mathcal{L}_{silog}}^{\mathcal{L}_{gt}} +\overbrace{\lambda_1\mathcal{L}_{rank} +\lambda_2\mathcal{L}_{ssi}}^{\mathcal{L}_{pl}},
\end{equation}

\noindent where $\lambda_1$ and $\lambda_2$ are two balancing factors, both empirically set to 0.1 in our experiments, respectively. The terms $\mathcal{L}_{gt}$ and $\mathcal{L}_{pl}$ represent the supervision signals derived from the ground truth and pseudo labels, respectively.

\section{Experiments}
\label{sec:experiments}

\begin{figure*}[t]
\setlength\tabcolsep{1pt}
\centering
\small
    \begin{tabular}{@{}*{4}{C{3.0cm}}@{}}
    \includegraphics[width=1\linewidth]{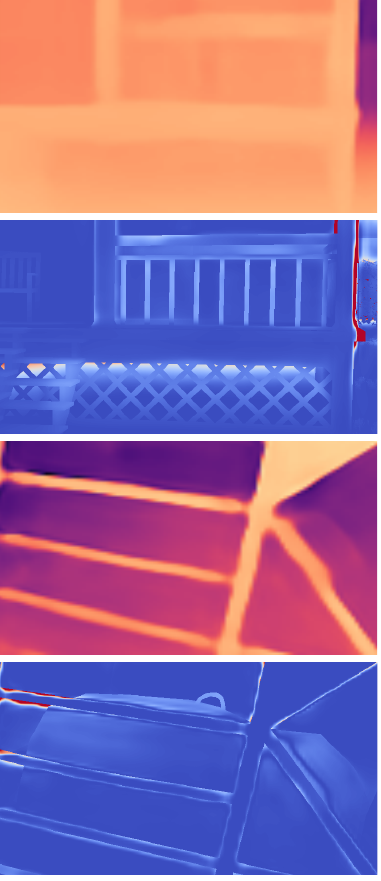} &
    \includegraphics[width=1\linewidth]{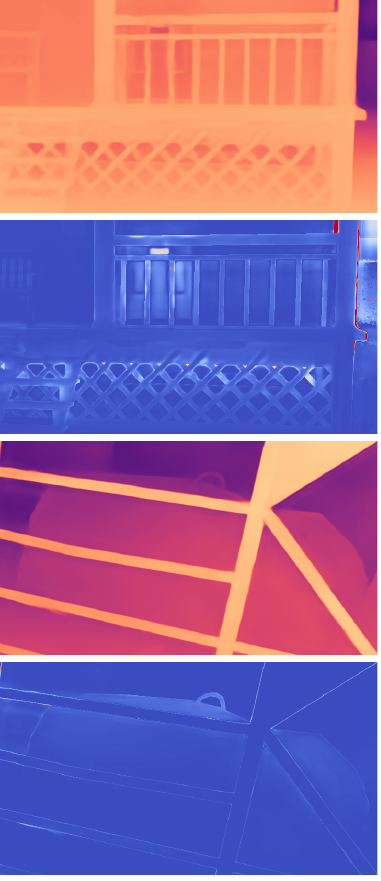} &
    \includegraphics[width=1\linewidth]{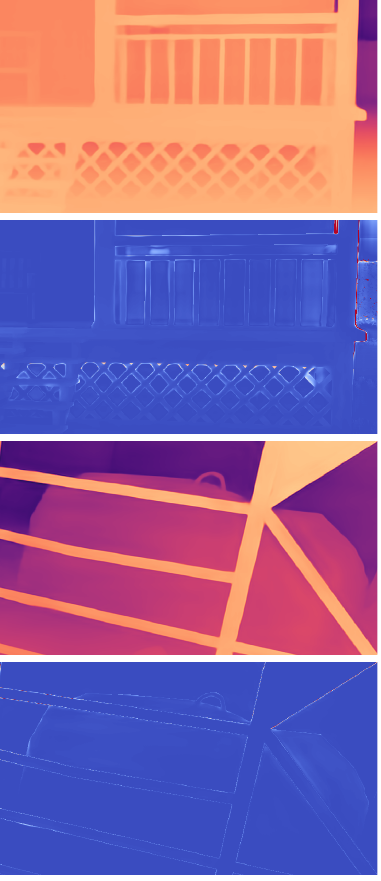} &
    \includegraphics[width=1\linewidth]{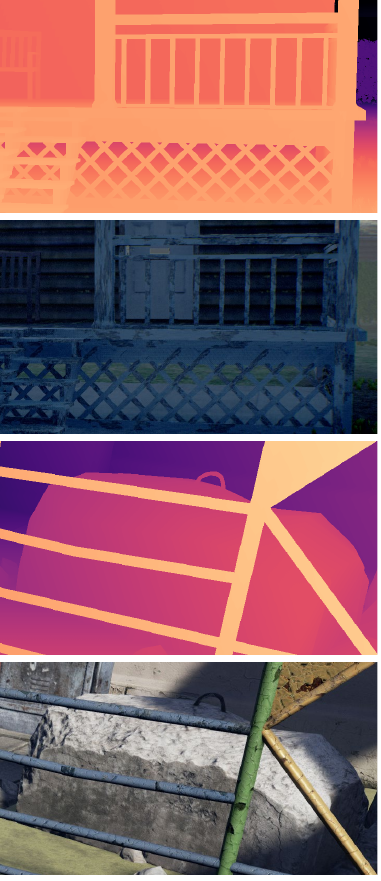} \\
    
    ZoeDepth~\cite{bhat2023zoedepth} & PatchFusion~\cite{li2023patchfusion} & Ours & GT, Input \\
    \end{tabular}
    \caption{\textbf{Qualitative Comparison on UnrealStereo4K.} We show the depth prediction and corresponding error map, respectively. The qualitative comparisons showcased here indicate our framework outperforms counterparts~\cite{bhat2023zoedepth,li2023patchfusion} with sharper edges and lower error around boundaries. We show individual patches in all images to emphasize details near depth boundaries.}
    \label{fig:u4k}
\end{figure*}

\subsection{Datasets and Metrics}

\noindent \textbf{UnrealStereo4K (Synthetic):} The UnrealStereo4K dataset~\cite{tosi2021smd} offers synthetic stereo images at a 4K resolution (2160$\times$3840), each paired with accurate, boundary-complete pixel-wise ground truth. Following the procedure in~\cite{li2023patchfusion}, we exclude mislabeled images using the Structural Similarity Index (SSIM)~\cite{wang2004ssim} and compute ground truth depth maps from provided disparity maps using camera parameters. Adhering to the dataset splits in~\cite{tosi2021smd,li2023patchfusion}, we employ a default patch size of 540$\times$960 for compatibility with~\cite{li2023patchfusion}.

\noindent \textbf{CityScapes (Real, Stereo):} Cityscapes~\cite{cordts2016cityscapes} offers a comprehensive suite of urban scene images, segmentation masks, and disparity maps at a resolution of 1024$\times$2048, surpassing many real-domain datasets in density, quantity, and resolution~\cite{silberman2012nyu,song2015sunrgbd,schops2017eth3d,scharstein2014mid}. We use a standard patch size of 256$\times$512 and conduct most experiments on this dataset.

\noindent \textbf{ScanNet++ (Real, LiDAR, Reconstruction):} ScanNet++~\cite{yeshwanth2023scannet++} is an extensive indoor dataset providing high-resolution images (1440$\times$1920), depth maps from iPhone LiDAR (192$\times$256), and high-resolution depth maps sampled from reconstructions of laser scans (1440$\times$1920). Our chosen patch size is 720$\times$960. We use the low-resolution ground-truth for training since the high-resolution version contains much noise as shown in Fig.~\ref{fig:dataset}.

\noindent \textbf{ETH3D (Real, LiDAR):} The ETH3D benchmark~\cite{schops2017eth3d} provides high-resolution indoor and outdoor images (6048$\times$4032) with ground-truth depth from laser sensors. We downsample the image-depth pairs to 2160$\times$3840 and select a patch size of 540$\times$960.

\noindent \textbf{Metrics:} We adopt standard depth evaluation metrics from~\cite{eigen2014mde,piccinelli2023idisc,bhat2023zoedepth} and the Soft Edge Error (SEE) from~\cite{tosi2021smd,chen2019over,li2023patchfusion} for \textit{scale} evaluation. Additional metric details are in the supplementary material. Given real-world depth map boundary incompleteness, we design metrics to focus on evaluating \textit{boundary} accuracy, leveraging high-resolution fine-grained segmentation masks as depth boundary quality proxies. Specifically, we use the Sobel operator~\cite{kanopoulos1988design} on both predicted depth and segmentation maps to generate edge masks, and then compute Precision, Recall, and F1 Score.

\begin{figure*}[t]
\setlength\tabcolsep{1pt}
\centering
\small
    \begin{tabular}{@{}*{5}{C{2.35cm}}@{}}
    \includegraphics[width=1\linewidth]{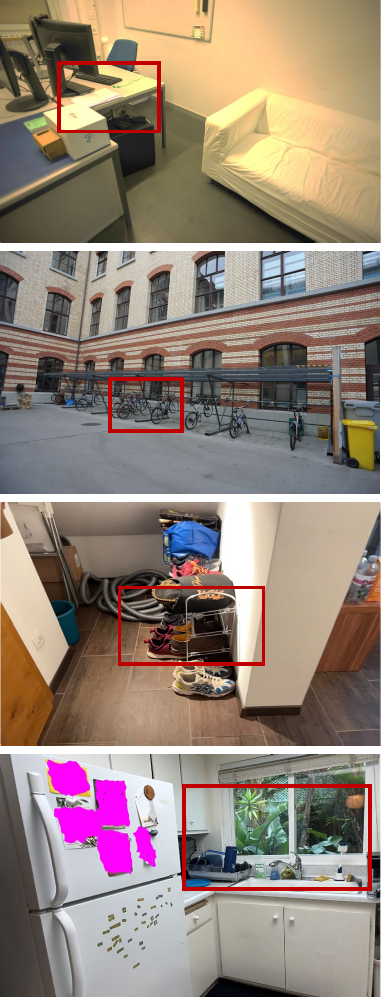} &
    \includegraphics[width=1\linewidth]{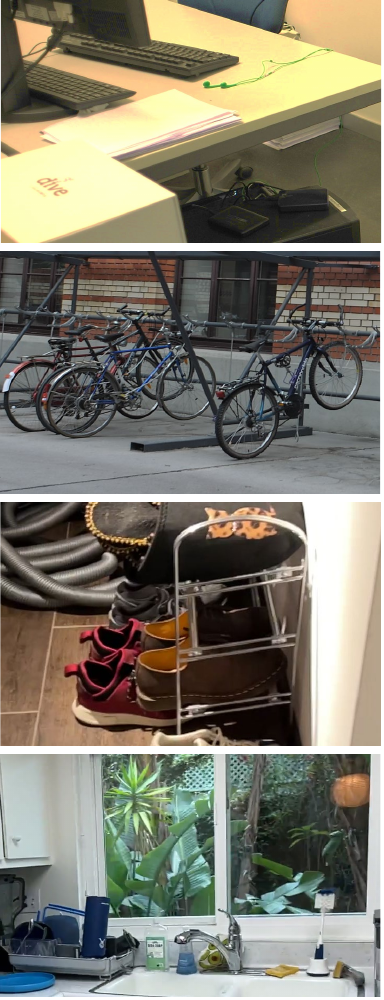} &
    \includegraphics[width=1\linewidth]{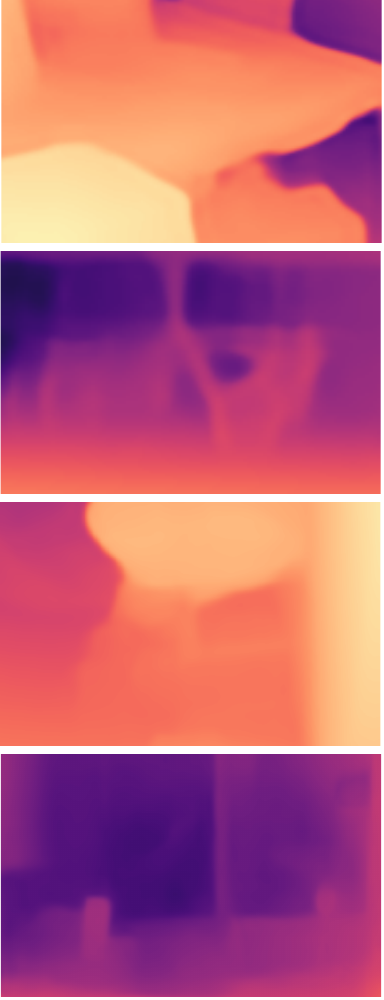} &
    \includegraphics[width=1\linewidth]{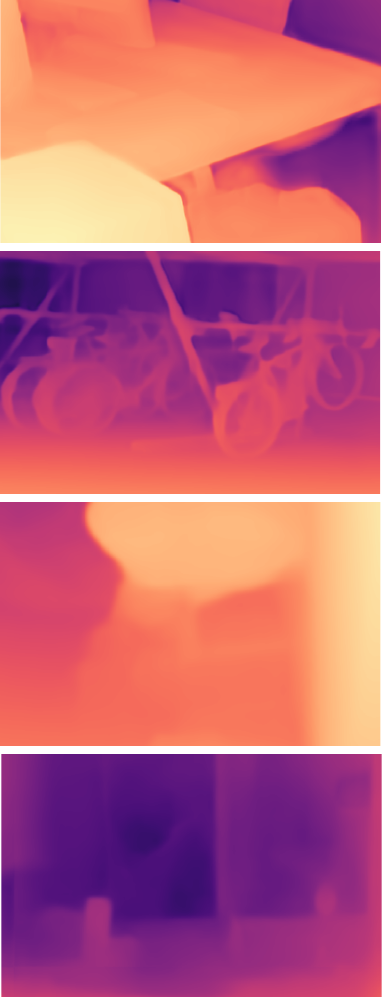} &
    \includegraphics[width=1\linewidth]{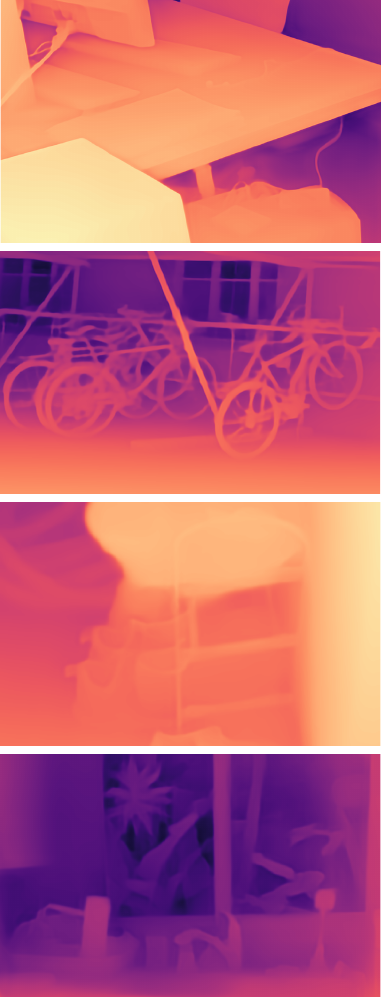} \\
    
    \multicolumn{2}{c}{Image} & ZoeDepth~\cite{bhat2023zoedepth} & PR $\mathcal{R}$ & Ours $1$\\
    \end{tabular}
    \caption{\textbf{Qualitative Comparison on ETH3D and ScanNet++.} The first two rows depict results for ETH3D and the last two for ScanNet++. The baseline ZoeDepth and PatchRefiner with conventional fine tuning (PR $\mathcal{R}$) both struggle to create high-resolution depth details near boundaries. Our proposed strategy yields crisp boundaries on both datasets.}
    \label{fig:ethscan}
\end{figure*}

\begin{figure*}[t]
\setlength\tabcolsep{1pt}
\centering
\small
    \begin{tabular}{@{}*{5}{C{2.35cm}}@{}}
    \includegraphics[width=1\linewidth]{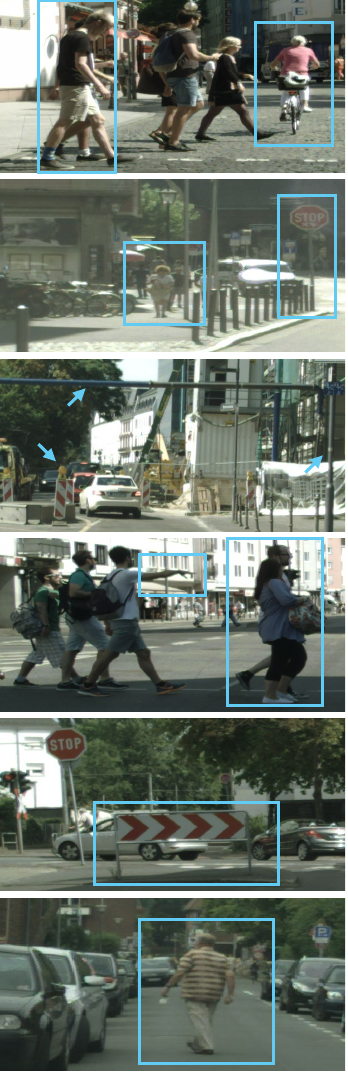} &
    \includegraphics[width=1\linewidth]{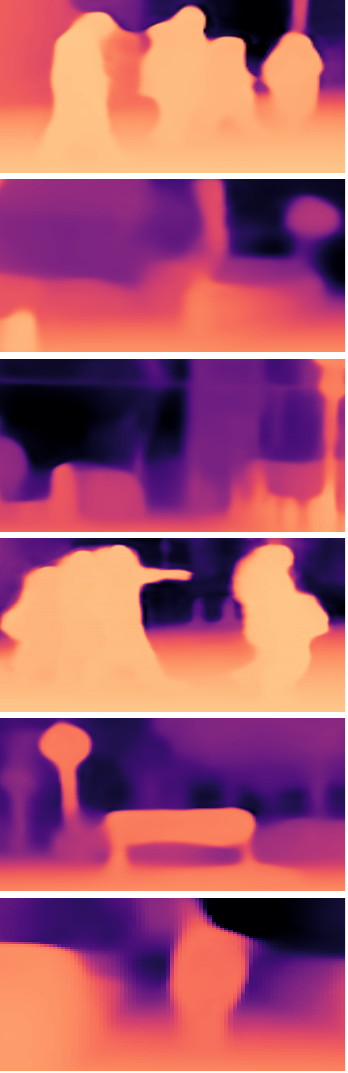} &
    \includegraphics[width=1\linewidth]{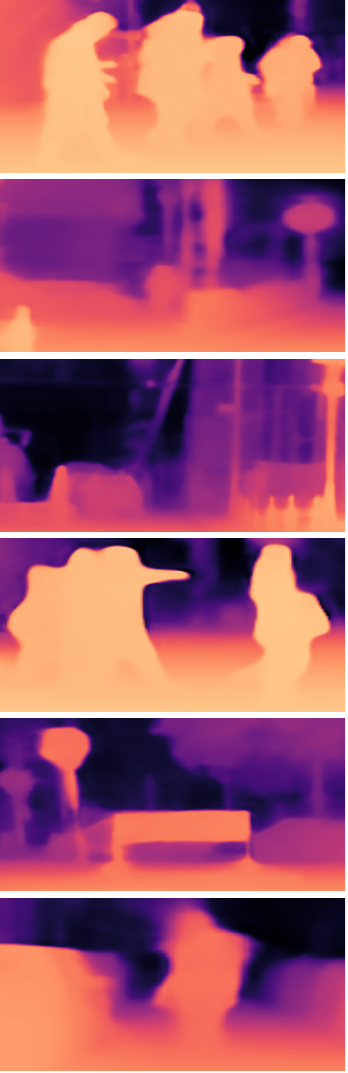} &
    \includegraphics[width=1\linewidth]{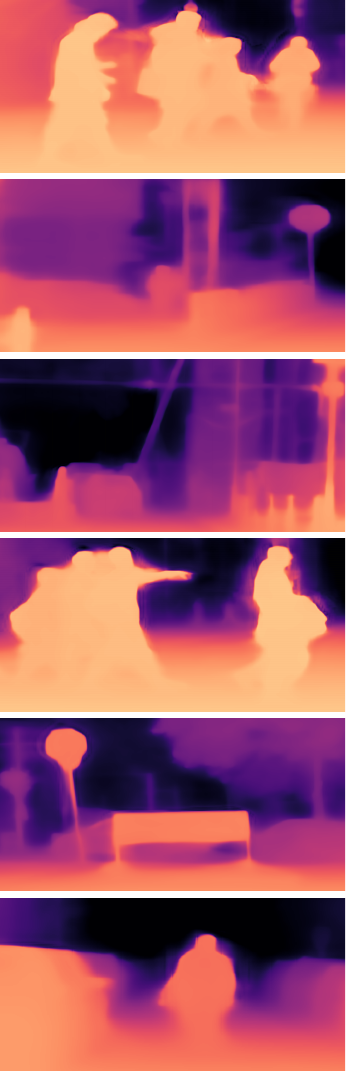} &
    \includegraphics[width=1\linewidth]{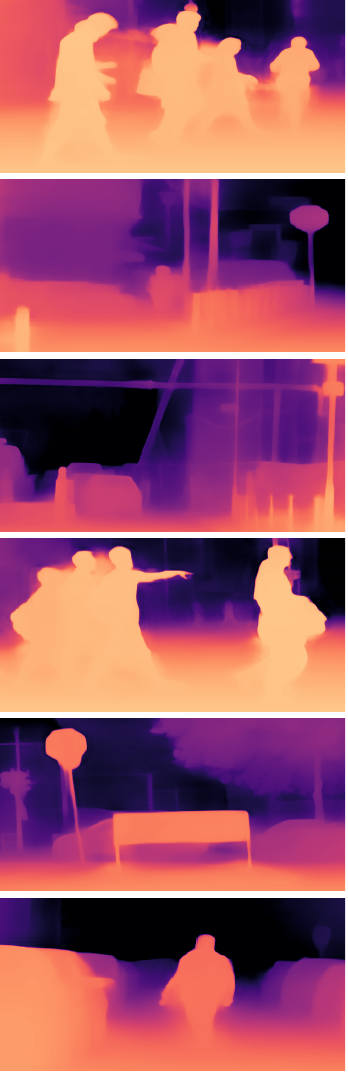} \\
    
    Image & ZoeDepth~\cite{bhat2023zoedepth} & PR $\mathcal{R}$ & Ours $1e^{-1}$ & Ours $1$\\
    \end{tabular}
    \caption{\textbf{Qualitative Comparison on CityScapes.} This figure illustrates depth estimation comparisons between the base ZoeDepth model, PatchRefiner (PR) trained on CityScapes, and our method. We display outcomes under varying levels of $\mathcal{L}_{pl}$ supervision ($\lambda_1=\lambda_2=1e^{-1}$ or $1$), featuring zoomed-in sections of each image to highlight detail fidelity near depth discontinuities.}
    \label{fig:cs}
\end{figure*}

\begin{table*}[t!]
    \centering
    \caption{\textbf{Quantitative comparison on UnrealStereo4K.} We color code the corresponding best competitor and our method within each block. PF and PR are short for PatchFusion~\cite{li2023patchfusion} and PatchRefiner, respectively. The reported numbers are from~\cite{li2023patchfusion}.}
    \label{tab:arch}
    \scalebox{0.8}{
    \begin{tabular}{l|*{5}{C{1.5cm}}|C{1.9cm}}
        \toprule
        Method & \boldsymbol{$\delta_1 (\%)$}$\uparrow$ & \textbf{REL}$\downarrow$ & \textbf{RMS}$\downarrow$ & \textbf{SiLog}$\downarrow$ & \textbf{SEE}$\downarrow$ & Reference \\
        \midrule
        iDisc~\cite{piccinelli2023idisc}          & 96.940 & 0.053 & 1.404 & 8.502 & 1.070  & ICCV 2023 \\
        SMD-Net~\cite{tosi2021smd} & 97.774 & 0.044 & 1.282 & 7.389 &  0.883 & CVPR 2021  \\
        Graph-GDSR~\cite{de2022gdsr} & 97.932 & 0.044 & 1.264 & 7.469 & 0.872 & CVPR 2022 \\
        BoostingDepth~\cite{miangoleh2021boostingdepth} & 98.104 & 0.044 & 1.123 & 6.662 & 0.939 & CVPR 2021 \\
        \midrule
        \midrule
        
        ZoeDepth~\cite{bhat2023zoedepth}  & 97.717 & 0.046 & 1.289 & 7.448 & 0.914 & - \\ 
        \hline
        ZoeDepth+PF$_{\textsc{P=16}}$~\cite{li2023patchfusion} &\cellcolor{Apricot} 98.419 &\cellcolor{Apricot} 0.040 &\cellcolor{Apricot} 1.088 &\cellcolor{Apricot} 6.212 &\cellcolor{Apricot} 0.838 & \multirow{3}{*}{CVPR 2024} \\
        ZoeDepth+PF$_{\textsc{P=49}}$~\cite{li2023patchfusion}  &\cellcolor{Salmon} 98.450 &\cellcolor{Salmon} 0.039 &\cellcolor{Salmon} 1.075 &\cellcolor{Salmon} 6.131 &\cellcolor{Salmon} 0.846  \\
        ZoeDepth+PF$_{\textsc{R=128}}$~\cite{li2023patchfusion}  &\cellcolor{Goldenrod} 98.469 &\cellcolor{Goldenrod} 0.039 &\cellcolor{Goldenrod} 1.066 &\cellcolor{Goldenrod} 6.085 &\cellcolor{Goldenrod} 0.849  \\
        \midrule
        ZoeDepth+\textbf{PR}$_{\textsc{P=16}}$ & \cellcolor{Apricot}98.821 & \cellcolor{Apricot}0.033 & \cellcolor{Apricot}0.892 & \cellcolor{Apricot}5.417 &  \cellcolor{Apricot}0.750 & \multirow{3}{*}{\textbf{Ours}} \\ 
        ZoeDepth+\textbf{PR}$_{\textsc{P=49}}$ &\cellcolor{Salmon} 98.859 &\cellcolor{Salmon} 0.033 &\cellcolor{Salmon}0.870 &\cellcolor{Salmon} 5.319 &\cellcolor{Salmon} 0.751  \\
        ZoeDepth+\textbf{PR}$_{\textsc{R=128}}$ &\cellcolor{Goldenrod}98.864 &\cellcolor{Goldenrod} 0.033 &\cellcolor{Goldenrod} 0.872 &\cellcolor{Goldenrod} 5.377 &\cellcolor{Goldenrod} 0.738  \\
        \midrule
        \midrule
        Depth-Anything~\cite{yang2024depthanything}                 & 97.773 & 0.041 & 1.235 & 7.192 & 0.911 & CVPR 2024 \\
        \midrule
        Depth-Anything+PF$_{\textsc{P=16}}$~\cite{li2023patchfusion} &\cellcolor{Apricot} 98.558 &\cellcolor{Apricot} 0.036 &\cellcolor{Apricot} 1.015 &\cellcolor{Apricot} 5.883 &\cellcolor{Apricot} 0.811 & \multirow{3}{*}{CVPR 2024} \\
        Depth-Anything+PF$_{\textsc{P=49}}$~\cite{li2023patchfusion} &\cellcolor{Salmon} 98.607 &\cellcolor{Salmon}0.035 &\cellcolor{Salmon} 0.987 &\cellcolor{Salmon} 5.746 &\cellcolor{Salmon} 0.812  \\
        Depth-Anything+PF$_{\textsc{R=128}}$~\cite{li2023patchfusion} &\cellcolor{Goldenrod} 98.616 &\cellcolor{Goldenrod}0.035 &\cellcolor{Goldenrod} 0.984 &\cellcolor{Goldenrod} 5.775 &\cellcolor{Goldenrod} 0.813  \\
        \midrule
        Depth-Anything+\textbf{PR}$_{\textsc{P=16}}$ &\cellcolor{Apricot} 98.826 &\cellcolor{Apricot} 0.033 &\cellcolor{Apricot} 0.889 &\cellcolor{Apricot} 5.289 &\cellcolor{Apricot} 0.768 & \multirow{3}{*}{\textbf{Ours}} \\
        Depth-Anything+\textbf{PR}$_{\textsc{P=49}}$ &\cellcolor{Salmon} 98.878 &\cellcolor{Salmon} 0.033 &\cellcolor{Salmon} 0.860 &\cellcolor{Salmon} 5.149 &\cellcolor{Salmon} 0.767  \\
        Depth-Anything+\textbf{PR}$_{\textsc{R=128}}$ &\cellcolor{Goldenrod} 98.878 &\cellcolor{Goldenrod}0.033 &\cellcolor{Goldenrod}0.860 &\cellcolor{Goldenrod} 5.206 &\cellcolor{Goldenrod} 0.759  \\
        \bottomrule
    \end{tabular}
    }
    
\end{table*}

\begin{table*}[t!]
    \centering
    \caption{\textbf{Quantitative comparison on CityScapes.} FT, and MIX are short for fine-tuning and mixed-data strategies, which are our main competitors. Baseline is highlighted in gray and each of the three competitors in a different color.}
    \label{tab:s2r}
    \scalebox{0.8}{
    \begin{tabular}{l|*{2}{C{0.5cm}}|*{3}{C{1.5cm}}|*{3}{C{1.5cm}}}
        \toprule
        \multirow{2}{*}{Method} & \multicolumn{2}{c|}{Data} & \multicolumn{3}{c|}{Scale} & \multicolumn{3}{c}{Boundary} \\
        \cmidrule{2-9}
        
        & $\mathcal{S}$ & $\mathcal{R}$ & \boldsymbol{$\delta_1 (\%)$}$\uparrow$ &  \textbf{REL}$\downarrow$ & \textbf{RMS}$\downarrow$  & \textbf{Precision}$\uparrow$ & \textbf{Recall}$\uparrow$ & \textbf{F1}$\uparrow$  \\
        \midrule
        ZoeDepth~\cite{bhat2023zoedepth} &  & \checkmark &  94.502 & 0.070 & 4.406 & 13.32 & 37.59 & 19.26  \\ 
        ZoeDepth~\cite{bhat2023zoedepth} + FT & \checkmark & \checkmark & 94.498 & 0.071 & 4.418 & 12.93 & 37.89 & 18.89 \\ 
        \hline
        PatchRefiner (zero-shot) & \checkmark &   & 5.705 & 0.399 & 12.203  & 28.68 & 51.28 & 36.34 \\
        PatchRefiner &   &  \checkmark & \cellcolor{lightgray} 95.284 & \cellcolor{lightgray} 0.066 & \cellcolor{lightgray} 4.047 & \cellcolor{lightgray} 16.67 & \cellcolor{lightgray} 39.53 & \cellcolor{lightgray} 23.04 \\
        PatchRefiner + FT & \checkmark & \checkmark &\cellcolor{Salmon} 95.418 &\cellcolor{Salmon} 0.065 &\cellcolor{Salmon} 3.992 &\cellcolor{Salmon} 17.09 &\cellcolor{Salmon} 40.92 &\cellcolor{Salmon} 23.68 \\ 
        PatchRefiner + mix~\cite{poucin2021boosting} & \checkmark & \checkmark &\cellcolor{Goldenrod} 89.108 &\cellcolor{Goldenrod} 0.112 &\cellcolor{Goldenrod} 4.732 &\cellcolor{Goldenrod} 23.08 &\cellcolor{Goldenrod} 41.17 &\cellcolor{Goldenrod}29.26 \\
        PatchRefiner + \textbf{DSD} & \checkmark & \checkmark & \cellcolor{Apricot}95.359 & \cellcolor{Apricot}0.066 & \cellcolor{Apricot}3.982  & \cellcolor{Apricot}18.92 & \cellcolor{Apricot}48.78 & \cellcolor{Apricot}26.84 \\ 
        \bottomrule
    \end{tabular}
    }
    
\end{table*}





 
 


\subsection{Implementation Details}

\noindent \textbf{PatchRefiner on Synthetic Dataset:}
For training on synthetic datasets, we employ the scale-invariant log loss $\mathcal{L}_{silog}$, as outlined in~\cite{eigen2014mde,bhat2023zoedepth,li2022binsformer,li2023patchfusion}. Initialization of the coarse network $\mathcal{N}_c$ leverages pretrained weights from the NYU-v2 dataset~\cite{silberman2012nyu}, adhering to the approach in~\cite{li2023patchfusion}. We dedicate 24 epochs to training $\mathcal{N}_c$, which is subsequently frozen to ensure stability in subsequent training phases. For the refinement network $\mathcal{N}_r$, initialization employs $\mathcal{N}_c$'s parameters, and training extends for an additional 36 epochs. Standard augmentation strategies from the baseline depth model are incorporated to enhance training effectiveness. During inference, we implement Consistency-Aware Inference, as described in~\cite{li2023patchfusion}, to optimize performance.

\noindent \textbf{Learning on Real-Domain Dataset:} Since the coarse model within the framework offers scale-consistent predictions, we first pretrain a coarse model on real-domain data to establish a reliable depth scale foundation. This model is subsequently frozen to preserve scale consistency throughout the training process. The refiner model in PatchRefiner is then initialized with parameters pretrained on synthetic data, enabling it to maintain high-frequency detail knowledge acquired from the synthetic domain. Initially, the student model is trained solely with $\mathcal{L}_{silog}$ for 24 epochs. Subsequent fine-tuning with our Detail and Scale Disentangling loss $\mathcal{L}_{DSD}$ over an additional 6 epochs to refine depth estimations.

\subsection{Main Results}

\noindent \textbf{Synthetic Dataset}: On the UnrealStereo4K dataset, PatchRefiner not only outperforms the base depth model but also shows substantial improvements over PatchFusion, reducing RMSE by 18.1\% and REL by 15.7\%. This advancement is further underscored by achieving the lowest SEE, highlighting our model's proficiency in capturing edge details. Qualitative comparisons, illustrated in Fig.~\ref{fig:u4k}, indicate the superior boundary delineation achieved by PatchRefiner.

\noindent \textbf{Real-Domain Dataset}: Tab.~\ref{tab:s2r} and Fig.~\ref{fig:ethscan},~\ref{fig:cs} delineate the performance disparity when leveraging synthetic data for real-domain learning. Although the synthetic-trained model excels in boundary details, it fails in scale accuracy due to the domain gap. Sole training on real-domain data enhances baseline's scale prediction yet falls short in detail accuracy due to the missing depth ground truth around boundaries. Neither fine-tuning nor mixed-training substantially elevates performance across scale and detail metrics, reflecting the inherent challenges in our task. Contrastingly, our strategy propels the model to notable gains in boundary accuracy (19.2\% increase in boundary recall) while sustaining scale precision comparable to the baseline model.

\begin{table*}[t!]
    \centering
    \caption{\textbf{Ablation study of architecture variations and formulation of final depth prediction}. Ours is highlighted with color. We highlight the best in \textbf{bold}.
    }
    \label{tab:abl-arch}
    \scalebox{0.78}{
    \begin{tabular}{l|C{2.3cm}|C{2.3cm}|*{5}{C{1.5cm}}}
        \toprule
        Method & Type of output & Feature Levels & \boldsymbol{$\delta_1 (\%)$}$\uparrow$ & \textbf{REL}$\downarrow$ & \textbf{RMS}$\downarrow$ & \textbf{SiLog}$\downarrow$ & \textbf{SEE}$\downarrow$  \\
        \midrule
        PatchFusion~\cite{li2023patchfusion} & Direct & 6 features & 98.419 & 0.040 & 1.088 & 6.212 & 0.838 \\
        \midrule
        \multirow{8}{*}{PatchRefiner}  & $\mathbf{D}_{c}$ residual & 1 features & 98.734 & 0.034 & 0.926 & 5.550 & 0.782 \\
          & $\mathbf{D}_{c}$ residual & 2 features & 98.814 & \textbf{0.033} & 0.905 & 5.511 & \textbf{0.750} \\
          & $\mathbf{D}_{c}$ residual & 3 features & 98.815 & 0.034 & 0.900 & 5.583 & 0.753 \\ 
          & $\mathbf{D}_{c}$ residual & 4 features & 98.815 & \textbf{0.033} & 0.899 & 5.494 & 0.752 \\
          & $\mathbf{D}_{c}$ residual & 5 features & 98.814 & \textbf{0.033} & 0.894 & 5.468 & 0.752 \\
          & \cellcolor{Goldenrod} $\mathbf{D}_{c}$ residual & \cellcolor{Goldenrod} {6 features} & \cellcolor{Goldenrod} \textbf{98.821} & \cellcolor{Goldenrod} \textbf{0.033} & \cellcolor{Goldenrod} \textbf{0.892} & \cellcolor{Goldenrod} \textbf{5.417} & \cellcolor{Goldenrod} \textbf{0.750} \\ 
          \cmidrule{2-8}
          & $\mathbf{D}_{d}$ residual & 6 features & 98.804 & \textbf{0.033} & 0.899 & 5.448 &  0.753 \\
          & Direct & 6 features & 98.749 & 0.034 & 0.925 &  5.591 & 0.765 \\
        \bottomrule
    \end{tabular}
    }
\end{table*}

\begin{table*}[t!]
    \centering
    \caption{\textbf{Variations of $\mathcal{L}_{pl}$}. We analyse various options for $\mathcal{L}_{pl}$ and compare them against PR. $\mathcal{S}$ and PR. $\mathcal{R}$ that serve as baselines for training on Synthetic and Real data individually. We set $\lambda=\lambda_1=\lambda_2$ to analyze the influence of DSD weight. The highlighted result is achieved with $\lambda=1e^{-1}$.}
    \label{tab:abl-s2r}
    \scalebox{0.82}{
    \begin{tabular}{r|*{3}{C{1.8cm}}|*{3}{C{1.8cm}}}
        \toprule
        \multirow{2}{*}{Variations ($\mathcal{L}_{pl}$)} & \multicolumn{3}{c|}{Scale} & \multicolumn{3}{c}{Boundary} \\
        \cmidrule{2-7}
        
        & \boldsymbol{$\delta_1 (\%)$}$\uparrow$ &  \textbf{REL}$\downarrow$ & \textbf{RMS}$\downarrow$  & \textbf{Precision}$\uparrow$ & \textbf{Recall}$\uparrow$ & \textbf{F1}$\uparrow$  \\
        \midrule
        Baseline PR. $\mathcal{S}$ & 5.705 & 0.399 & 12.20 & 28.68 & 51.28 & 36.34 \\ 
        Baseline PR. $\mathcal{R}$ & 95.418 & 0.065 & 3.992 & 17.09 & 40.92 & 23.68 \\ 
        \midrule
        $\mathcal{L}_{silog}$ & 82.550 & 0.155 & 5.914 & 26.34 & 58.35 & 35.81\\ 
        $\mathcal{L}_{silog}$ + mask & 81.425 & 0.148 & 6.513 & 30.04 & 46.89 & 36.19\\ 
        $\mathcal{L}_{rank}$ & 95.413 & 0.066 & 3.973 & 18.38 & 47.81 & 26.12 \\ 
        $\mathcal{L}_{ssi}$ & 95.465 & 0.065 & 3.974 & 19.26 & 44.01 & 26.39 \\ 
        $\mathcal{L}_{rank}$ + $\mathcal{L}_{ssi}$ (Ours) & \cellcolor{Goldenrod} 95.359 & \cellcolor{Goldenrod} 0.066 & \cellcolor{Goldenrod} 3.982 & \cellcolor{Goldenrod} 18.92 & \cellcolor{Goldenrod} 48.78 & \cellcolor{Goldenrod} 26.84 \\ 

        \midrule
        $\lambda=1$ & 95.077 & 0.069 & 4.222 & 21.32 & 58.48 & 30.90 \\ 
        $\lambda=3e$$^{-1}$ & 95.296 & 0.068 & 4.086 &  20.11 &  53.83 & 28.91 \\ 
        $\lambda=3e$$^{-2}$ & 95.462 & 0.065 & 3.953 & 18.01 & 43.90 & 25.11 \\ 
        \bottomrule
    \end{tabular}
    }
\vspace{-0.15cm}
\end{table*}

\subsection{Ablation Studies and Discussion}
\label{subsec:ablation}

We ablate and discuss the contributions of individual components. As default, we utilize the UnrealStereo4K dataset for synthetic-dataset training and the PatchRefiner variant with $P=16$ patches for clarity and ease of comparison. We ablate our teacher-student framework using the CityScapes dataset.

\subsubsection{Architecture and Formulation Variations:}

We first ablate the effectiveness of our architecture design. Utilizing a ZoeDepth model, we extract six levels of intermediate features during the forward pass, represented as $\mathcal{F} = {f_1, f_2,\cdots,f_6}$, with progressively increasing resolution. We then sequentially omit lower resolution feature maps from the refiner's decoder input. As depicted in Tab.~\ref{tab:abl-arch}, performance diminishes with fewer feature maps, yet even with solely the highest resolution feature map, our model outperforms PatchFusion.

Further, we explore our refinement formulation's efficacy by comparing it against two alternatives: (1) Residual depth adjustment based on the base model $\mathcal{N}_d$ within the refiner. (2) Direct depth prediction from the refiner's decoder, similar to PatchFusion ($\mathbf{D} = \mathbf{D}_r$ in this context). Our residual-based approach demonstrates superior performance, underscoring its advantages in optimization and training efficiency.

\subsubsection{Effectiveness of Pseudo Label Supervision:}

We evaluate the efficacy of our Detail and Scale Disentangling (DSD) loss against the conventional scale-invariant loss, $\mathcal{L}_{silog}$, for pseudo label supervision. Tab.~\ref{tab:abl-s2r} illustrates that while $\mathcal{L}_{silog}$ aids in detail transfer from the teacher to the student model, it compromises scale accuracy due to significant discrepancies in the pseudo labels. A masking approach, focusing only on areas lacking depth, does not mitigate this issue, indicating pervasive negative effects. In contrast, the combination of $\mathcal{L}_{rank}$ and $\mathcal{L}_{ssi}$ within $\mathcal{L}_{pl}$ not only improves detail fidelity but also maintains scale accuracy, demonstrating that ranking constraints and scale-shift invariance are effective, orthogonal strategies for enhancing high-resolution detail without sacrificing scale accuracy.

\subsubsection{Exploration of Variant DSD Loss Weight:}

This study examines the impact of different DSD loss weights ($\lambda_1$ and $\lambda_2$) to elucidate the loss's efficiency. As shown in Tab.~\ref{tab:abl-s2r} and Fig.~\ref{fig:cs}, increasing the loss weight marginally affects scale accuracy while significantly improving boundaries, validating the DSD loss's role in balancing detail enhancement and scale preservation. Notably, when both weights are set to $1$, the model's boundary recall surpasses the teacher's performance. Furthermore, when DSD loss achieves a comparable boundary metric to $\mathcal{L}_{silog}$, it exhibits a smaller performance decline, underscoring its effectiveness.
\section{Conclusion}
\label{sec:conclusion}

We presented \textbf{PatchRefiner}, a tile-based framework tailored to real-world high-resolution monocular metric depth estimation. It reconceptualizes high-resolution depth estimation as a refinement process.
With a pseudo-labeling strategy that leverages synthetic data, we propose a Detail and Scale Disentangling (DSD) loss to enhance detail capture while maintaining scale accuracy. Our proposed framework decisively surpasses the current SOTA method for UnrealStereo4K (17.3\% in RMSE), while demonstrating marked improvements in detail accuracy on real-world datasets such as CityScape, ScanNet++, and ETH3D.


%
%
\bibliographystyle{splncs04}
\bibliography{main}

\appendix

\appendix
\section{Boundary Evaluation Protocol}

This section addresses the challenges of directly applying traditional boundary discontinuity metrics~\cite{koch2018evaluation,spencer2023mdec1,spencer2023mdec2} to high-resolution depth estimation. Then, we introduce our approach, which utilizes high-resolution segmentation masks as proxies for assessing depth boundary quality. Our method computes the precision, recall, and F-1 Score metrics introduced in our paper and a refined depth boundary error metric detailed here. Together, this enables a comprehensive boundary evaluation on real-domain datasets.

Depth Boundary Error (DBE) is a standard metric for assessing depth discontinuity accuracy~\cite{koch2018evaluation,spencer2023mdec1,spencer2023mdec2}. It involves extracting boundary masks $\Tilde{\mathbf{M}}$ and $\mathbf{M}$ from the ground-truth (GT) and predicted depth maps, $\Tilde{\mathbf{D}}$ and $\mathbf{D}$, using methods like structured edges~\cite{dollar2014fast,koch2018evaluation}, Sobel~\cite{kanopoulos1988design,spencer2023mdec1,spencer2023mdec2}, or Canny operators~\cite{canny1986computational,spencer2023mdec1,spencer2023mdec2}. DBE calculates the \textit{truncated chamfer distance} between these masks, employing an \textit{Euclidean Distance Transform} (EDT) on edge maps while disregarding distances beyond a threshold $\theta$ to focus on the local edge evaluation. The accuracy measure $\varepsilon^{acc}_{DBE}$ and the completeness error $\varepsilon^{comp}_{DBE}$ quantify the proximity of predicted boundaries to the ground truth and vice versa, respectively. They are defined as

\begin{equation}
    \varepsilon^{acc}_{DBE} = \sum\mathrm{EDT}(\Tilde{\mathbf{M}}(p))\cdot\mathbf{M}(p),
\end{equation}
\begin{equation}
    \varepsilon^{comp}_{DBE} = \sum\mathrm{EDT}(\mathbf{M}(p))\cdot\Tilde{\mathbf{M}}(p).
\end{equation}

These metrics were proposed as part of the IBims-1~\cite{koch2018evaluation} benchmark, and recently adopted in monocular depth estimation challenges~\cite{spencer2023mdec1,spencer2023mdec2}. We refer to these papers for more details.

\begin{figure}[t]
    \centering
    \includegraphics[width=1\linewidth]{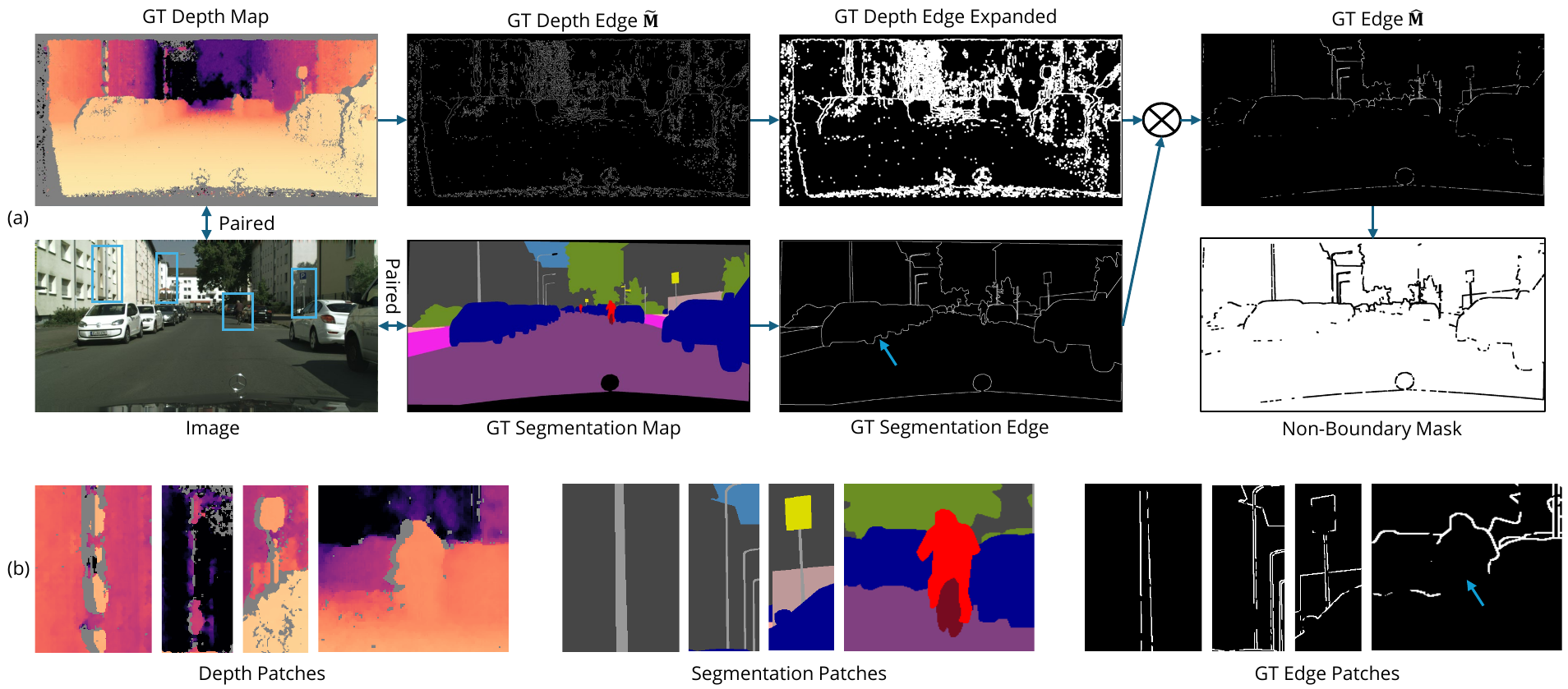}
    \caption{\textbf{Evaluation Pipeline and Noise in GT Depth Maps.} (a) We combine the information from GT segmentation maps and GT depth maps to obtain higher quality depth edges for the evaluation. (b) We showcase incorrectly labeled areas in the depth map, which influences the \textit{scale} evaluation. $\otimes$ denotes the pixel-wise \textit{and} operator.}
    \label{fig:eval}
\end{figure}

However, directly computing edge information on GT depth maps is impractical due missing values in the GT depth map. These missing values often occur close to edges, as shown in Fig.~\ref{fig:eval} (samples from CityScapes~\cite{cordts2016cityscapes} obtained by stereo GT system). Due to the missing values it is not possible to locate where exactly the edge is. Moreover, this issue also affects depth maps from LiDAR or scene reconstructions, characterized by sparsity and missing data (\eg, ETH3D~\cite{schops2017eth3d}, KITTI~\cite{geiger2012kitti} and high-resolution depth in ScanNet++~\cite{yeshwanth2023scannet++}). On the other hand, low-resolution depth maps are naturally not precise enough on boundaries for high-resolution depth prediction evaluation (\eg, low-resolution depth in ScanNet++~\cite{yeshwanth2023scannet++} and NYU~\cite{silberman2012nyu}).

To address these limitations, our main idea is to combine the information in GT depth maps with the information in GT segmentation maps.
We leverage segmentation maps as depth discontinuity indicators as follows. As shown in Fig.~\ref{fig:eval}, although these maps are noise-free, they include \textit{fake} edges not present in depth maps. We filter these edges using an expanded GT depth edge map, resulting in an accurate edge map, $\hat{\textbf{M}}$.

We follow the implementation of the monocular depth estimation challenge\footnote{\url{https://github.com/jspenmar/monodepth_benchmark}}\cite{spencer2023mdec1,spencer2023mdec2} to calculate the $\varepsilon^{comp}_{DBE}$ and $\varepsilon^{comp}_{DBE}$. To calculate the expanded GT depth edge map, we utilize a Gaussian blur with kernel size $k=7$. All pixels with value $>1$ are set as 1 as the expanded edge after the gaussian blur. The precision, recall, and F-1 Score used in our paper can be calculated with the depth prediction edge $\mathbf{M}$ and the final GT edge map $\hat{\textbf{M}}$. We present the results in Tab.~\ref{tab:sup-s2r}. Our approach achieves significant improvement on both of these metrics.

\begin{figure}[t]
    \centering
    \includegraphics[width=1\linewidth]{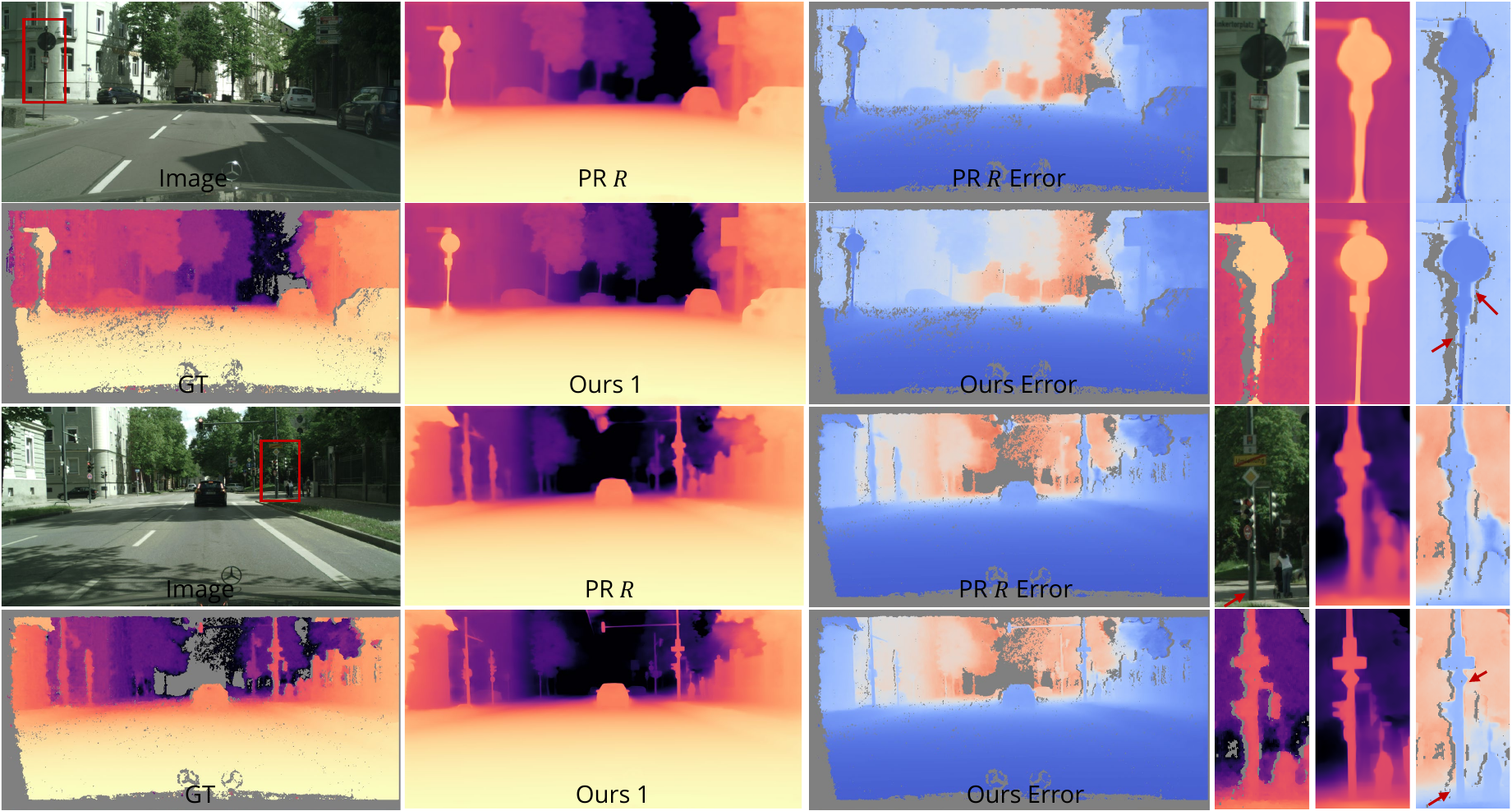}
    \caption{\textbf{Noise in GT Depth Maps.} While we achieve sharper boundaries that align with the input images, the incorrectly labeled pixels around the boundary on GT depth maps lead to an unreliable evaluation in \textit{scale} metrics. PR $\mathcal{R}$ denotes PatchRefiner with conventional fine tuning. While our results are much better, this improvement is not measurable with noisy GT depth maps.}
    \label{fig:mistake}
\end{figure}

\begin{table*}[t!]
    \centering
    \caption{\textbf{Quantitative Comparison on CityScapes}. Scale-paper and Scale-sup indicate the \textit{scale} evaluation on all valid pixels in main paper and on non-boundary pixel presented in supplementary materials, respectively.}
    \label{tab:sup-s2r}
    \scalebox{0.82}{
    \begin{tabular}{l|*{2}{C{1.8cm}}|*{2}{C{1.8cm}}|*{2}{C{1.8cm}}}
        \toprule
        \multirow{2}{*}{Method} & \multicolumn{2}{c|}{Scale-paper} & \multicolumn{2}{c|}{Scale-sup} & \multicolumn{2}{c}{Boundary} \\
        \cmidrule{2-7}
        
        & \boldsymbol{$\delta_1 (\%)$}$\uparrow$ & \textbf{RMS}$\downarrow$ & \boldsymbol{$\delta_1 (\%)$}$\uparrow$ & \textbf{RMS}$\downarrow$ &$\varepsilon^{acc}_{DBE}$$\downarrow$ &$\varepsilon^{comp}_{DBE}$$\downarrow$  \\
        \midrule
        PR $\mathcal{R}$ FT & 95.418 & 3.992 & 96.197 & 3.601 & 3.301 & 1.947 \\ 
        \midrule
        Ours $\lambda=1e^{-1}$ & 95.359 & 3.982 & 96.235 & 3.589 & 2.848 & 1.790 \\ 
        Ours $\lambda=1$ & 95.077 & 4.222 & 96.063 & 3.689 & 2.494 & 1.714 \\ 
        \bottomrule
    \end{tabular}
    }
\vspace{-0.15cm}
\end{table*}

\section{Challenges in Scale Evaluation}

In our primary paper, we assess the scale metric using all valid ground truth (GT) depth values to verify that our methods preserve the model's scale accuracy, adhering to the evaluation protocols established in prior monocular metric depth estimation research~\cite{eigen2014mde,bhat2021adabins,li2022binsformer,li2023patchfusion}. Specifically, we employ the root mean squared error (RMSE) $=|\frac{1}{M}\sum_{i=1}^M|d_i-\Tilde{d}_i|^2|^\frac{1}{2}$, mean absolute relative error (AbsRel) $=\frac{1}{M}\sum_{i=1}^M|d_i-\Tilde{d}_i|/d_i$, scale-invariant logarithmic error (SILog) $=|\frac{1}{M}\sum_{i=1}^Me^2-|\frac{1}{M}\sum_{i=1}^Me|^2|^{1/2} \times 100$ where $e=\log{\Tilde{d}_i} - \log{d_i}$, the average $\mathrm{log}_{10}$ error $=\frac{1}{M}\sum_{i=1}^M|\log_{10}d_i-\log_{10}\Tilde{d}_i|$ and the accuracy under the threshold ($\delta_i < 1.25^i, i = 1$), where $d_i$ and $\Tilde{d}_i$ refer to ground truth and predicted depth at pixel $i$, respectively, and $M$ is the total number of pixels in the image. 

However, as Fig.~\ref{fig:mistake} illustrates, GT depth values near edges exhibit higher errors, potentially skewing the scale metrics. To more accurately assess our method's effectiveness, this supplementary section presents scale metrics for non-boundary regions. For implementation, we re-use the final GT edge map $\hat{\textbf{M}}$ from the \textit{boundary} metric and apply an additional Gaussian blur (kernel size $k = 7$) to it, as shown in Fig.~\ref{fig:eval}. The comparative results are displayed in Tab.~\ref{tab:sup-s2r}, indicating the effectiveness of our method on maintaining the scale accuracy.

\begin{table*}[t!]
    \centering
    \caption{\textbf{Quantitative Comparison on ETH3D and ScanNet++.} We present the \textit{scale} metrics.}
    \label{tab:sup-ethcs}
    \scalebox{0.82}{
    \begin{tabular}{l|*{3}{C{1.8cm}}|*{3}{C{1.8cm}}}
        \toprule
        \multirow{3}{*}{Methods} & \multicolumn{3}{c|}{ETH3D} & \multicolumn{3}{c}{ScanNet++} \\
        \cmidrule{2-7}
        & \textbf{REL}$\downarrow$ & \textbf{RMS}$\downarrow$ & $\mathrm{\textbf{log}}_{10}$$\downarrow$ & \textbf{REL}$\downarrow$ & \textbf{RMS}$\downarrow$ & $\mathrm{\textbf{log}}_{10}$$\downarrow$ \\
        \midrule
        PR $\mathcal{R}$ FT & 0.147 & 1.431 & 0.061 & 0.145 & 0.268 & 0.059 \\ 
        \midrule
        Ours & 0.145 & 1.368 & 0.060 & 0.145 & 0.268 &  0.059 \\
        \bottomrule
    \end{tabular}
    }
\vspace{-0.15cm}
\end{table*}

\section{More Results}

\begin{figure*}[t]
\setlength\tabcolsep{1pt}
\centering
\small
    \begin{tabular}{@{}*{3}{C{3.8cm}}@{}}
    \includegraphics[width=1\linewidth]{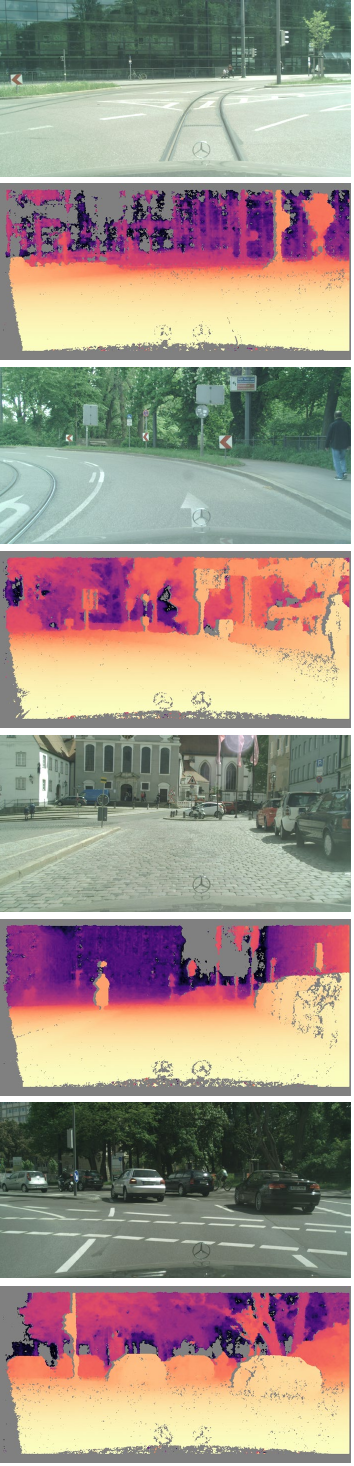} &
    \includegraphics[width=1\linewidth]{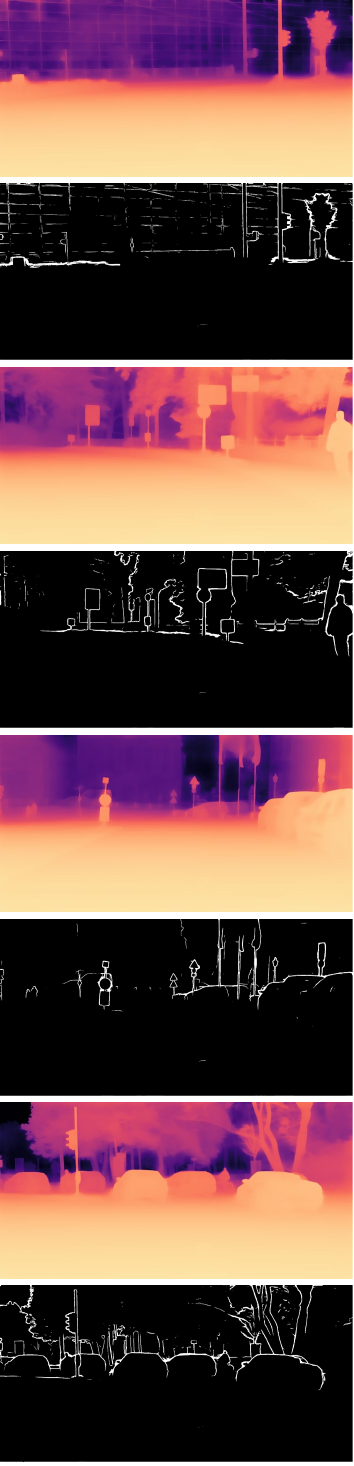} &
    \includegraphics[width=1\linewidth]{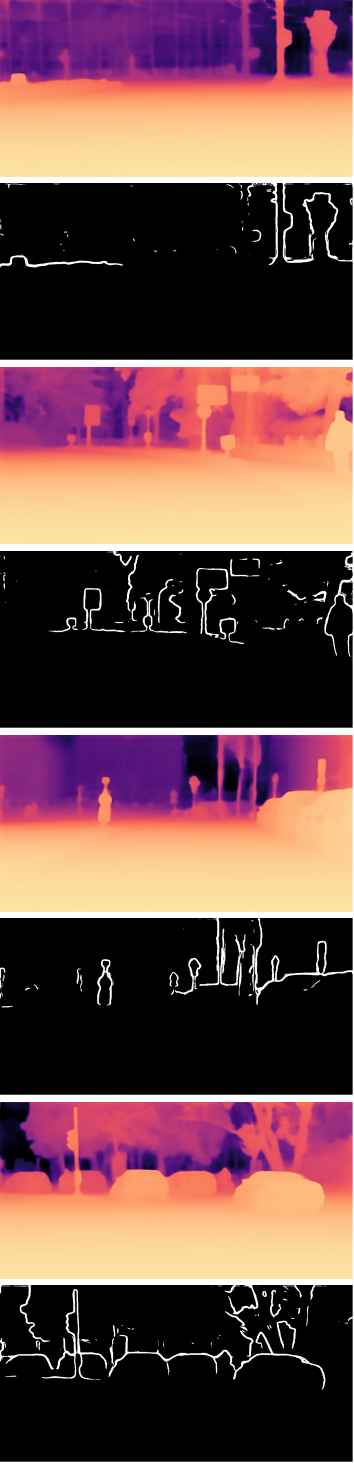} \\
    
    Input,GT & Ours & PR $\mathcal{R}$ FT \\
    \end{tabular}
    \caption{\textbf{Qualitative Comparison on CityScapes++.} We also present the boundary maps to show the effectiveness of our proposed method. PR $\mathcal{R}$ FT denotes PatchRefiner with conventional fine tuning. The resolution is 1024$\times$2048.}
    \label{fig:sup-cs}
\end{figure*}

\begin{figure*}[t]
\setlength\tabcolsep{1pt}
\centering
\small
    \begin{tabular}{@{}*{3}{C{3.8cm}}@{}}
    \includegraphics[width=1\linewidth]{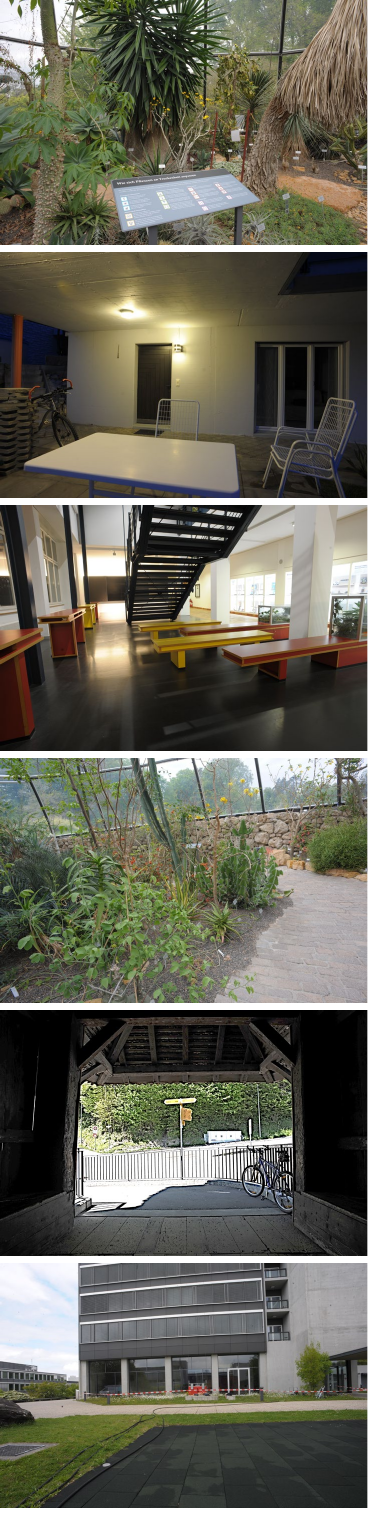} &
    \includegraphics[width=1\linewidth]{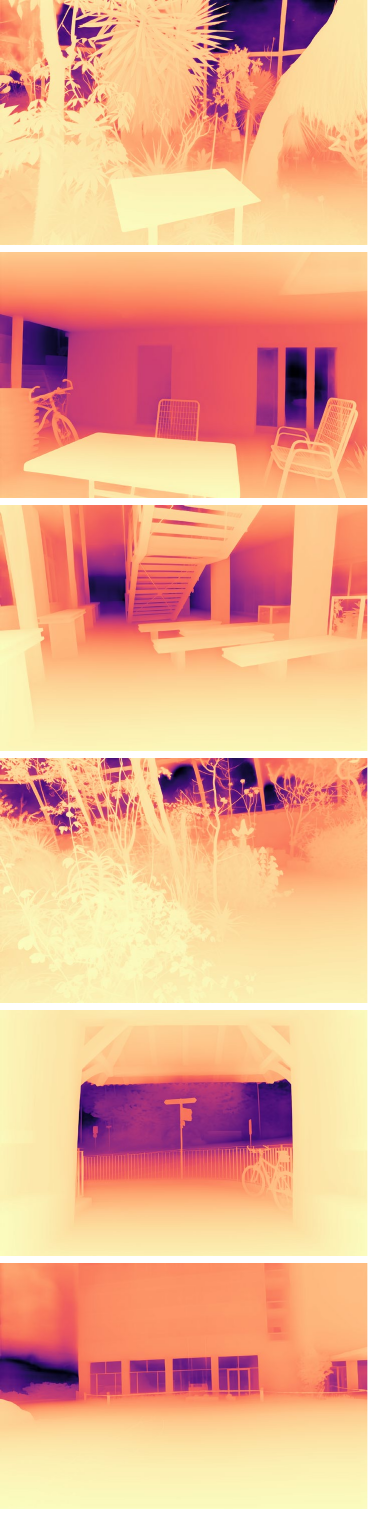} &
    \includegraphics[width=1\linewidth]{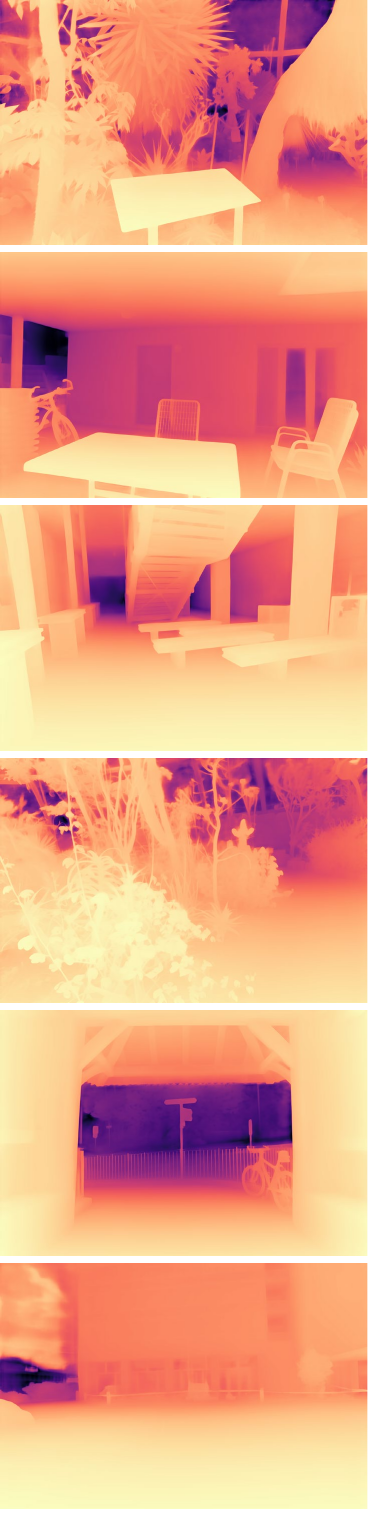} \\
    
    Input & Ours & PR $\mathcal{R}$ FT \\
    \end{tabular}
    \caption{\textbf{Qualitative Comparison on ETH3D.} PR $\mathcal{R}$ FT denotes PatchRefiner with conventional fine tuning. The resolution is 2160$\times$3840.}
    \label{fig:sup-eth}
\end{figure*}

\begin{figure*}[t]
\setlength\tabcolsep{1pt}
\centering
\small
    \begin{tabular}{@{}*{3}{C{3.8cm}}@{}}
    \includegraphics[width=1\linewidth]{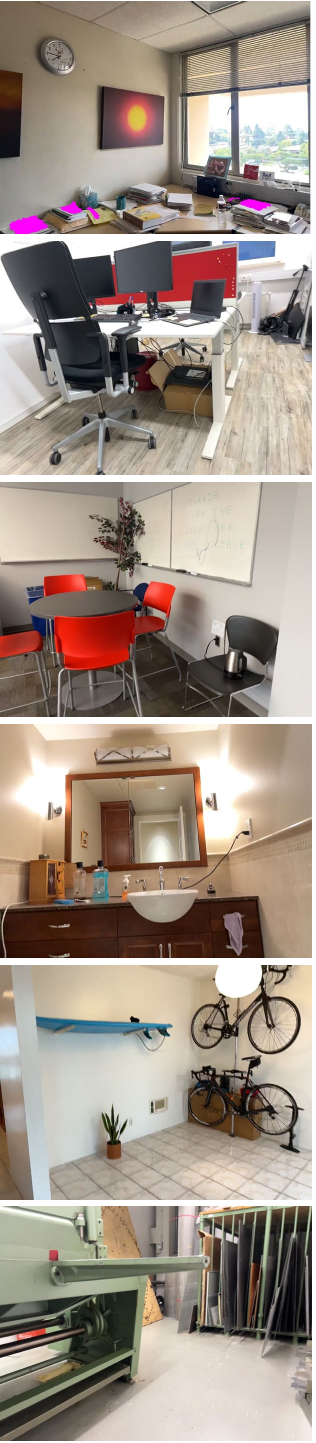} &
    \includegraphics[width=1\linewidth]{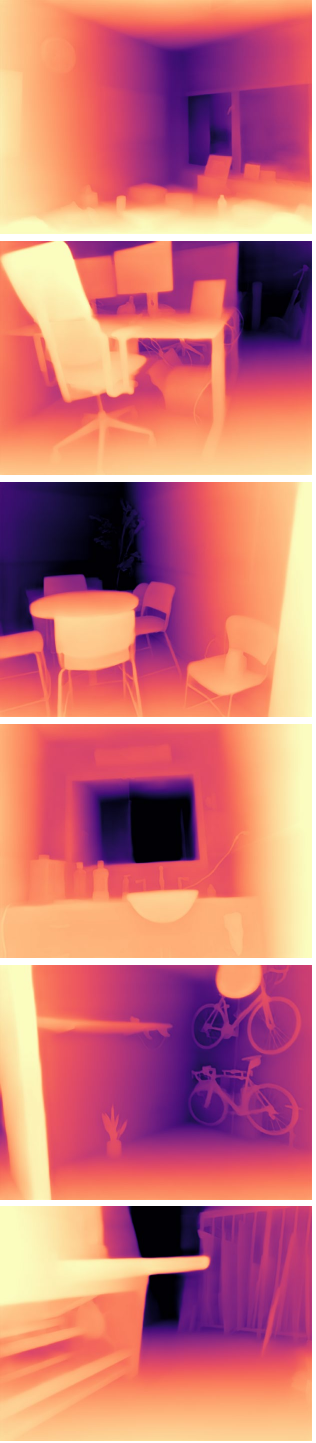} &
    \includegraphics[width=1\linewidth]{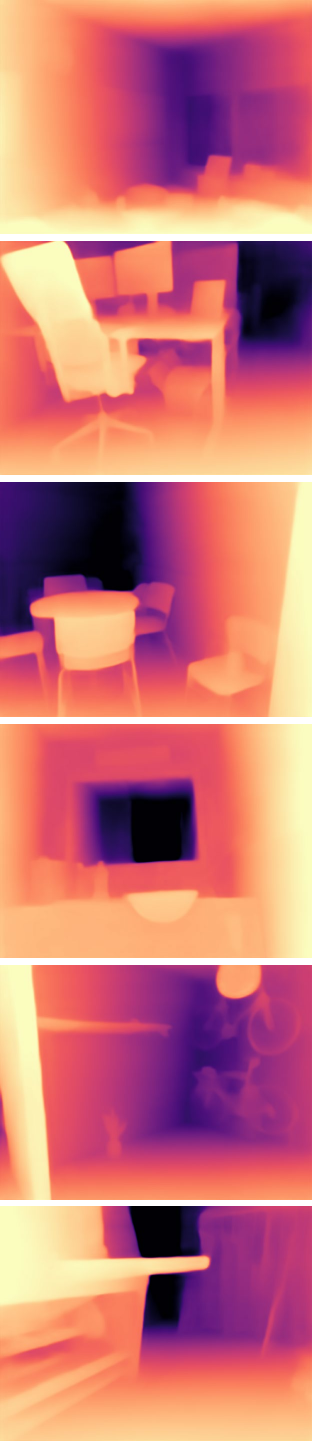} \\
    
    Input & Ours & PR $\mathcal{R}$ FT \\
    \end{tabular}
    \caption{\textbf{Qualitative Comparison on ScanNet++.} PR $\mathcal{R}$ FT denotes PatchRefiner with conventional fine tuning. The resolution is 1440$\times$1920.}
    \label{fig:sup-scannet}
\end{figure*}

We present additional qualitative results for CityScapes (Fig.~\ref{fig:sup-cs}), ETH3D (Fig.~\ref{fig:sup-eth}), and ScanNet++ (Fig.~\ref{fig:sup-scannet}). We also present \textit{scale} quantitative comparisons on ETH3D and ScanNet++ in Tab.~\ref{tab:sup-ethcs}. Here we calculate the result based on all valid GT pixels as described in our main paper. We present the results from \textit{Ours} $\lambda=1$ on the CityScapes dataset. We use $\lambda=5$ and $\lambda=3$ to train our model on ETH3D and ScanNet++, respectively. Coincidentally, the metrics on ScanNet++ are exactly the same.

\end{document}